# A strawberry harvest-aiding system with crop-transport co-robots: Design, development, and field evaluation

Chen Peng[1], Stavros Vougioukas[1], David Slaughter[1], Zhenghao Fei[1], Rajkishan Arikapudi[1]

[1]University of California, Davis, Department of Biological and Agricultural Engineering

## Abstract

Mechanizing the manual harvesting of fresh market fruits constitutes one of the biggest challenges to the sustainability of the fruit industry. During manual harvesting of some fresh-market crops like strawberries and table grapes, pickers spend significant amounts of time walking to carry full trays to a collection station at the edge of the field. A step toward increasing harvest automation for such crops is to deploy harvest-aid collaborative robots (co-bots) that transport the empty and full trays, thus increasing harvest efficiency by reducing pickers' non-productive walking times. This work presents the development of a co-robotic harvest-aid system and its evaluation during commercial strawberry harvesting. At the heart of the system lies a predictive stochastic scheduling algorithm that minimizes the expected non-picking time, thus maximizing the harvest efficiency. During the evaluation experiments, the co-robots improved the mean harvesting efficiency by around 10% and reduced the mean non-productive time by 60%, when the robot-to-picker ratio was 1:3. The concepts developed in this work can be applied to robotic harvest-aids for other manually harvested crops that involve walking for crop transportation.

# 1. Introduction

Mechanizing the manual harvesting of fresh market fruits constitutes one of the biggest challenges to the sustainability of the fruit industry. Depending on the commodity, labor for manual harvesting can contribute up to 60% of the yearly operating costs per acre (Bolda et al., 2016). Additionally, recent studies indicate that the farm labor supply cannot meet demand in many parts of the world because of socioeconomic, structural, and political factors (Charlton et al., 2019; Guan et al., 2015). Robotic harvester prototypes are being developed and field-tested for high-volume, high-value crops such as apples (Silwal et al., 2017), kiwifruit (Williams et al., 2020), sweet pepper (Arad et al., 2020), and strawberries (Xiong et al., 2020). However, the developed robots have not, to date, successfully replaced the judgment, dexterity, and speed of experienced pickers at a competing cost; the challenges of high fruit picking efficiency and throughput remain largely unsolved (Bac et al., 2014).

As an intermediate step to full automation, mechanical labor aids have been introduced to increase worker productivity by reducing workers' non-productive times. For example, orchard platforms eliminate the need for climbing ladders and walking to unload fruits in bins (Baugher et al., 2009; Fei & Vougioukas, 2021). Autonomous vehicle prototypes have been developed to assist in bin management in orchards (Bayar et al., 2015; Ye et al., 2017), to reduce the need for forklift operators.

In strawberry production, mobile conveyors have been introduced to reduce the time pickers spend walking to get the produce from the plants to the designated loading stations and return to resume picking (Rosenberg, 2003). However, such conveyors are specific to strawberries and cannot be adapted to other crops. Furthermore, their adoption has been very slow, partly because of their questionable profitability, due to high purchase cost and limited efficiency gains. Two reasons for their inadequate efficiency are: 1) row-turning in the field is time-consuming because of their large size, and 2) because conveyors move slowly to accommodate slower pickers, often resulting in underutilization of faster pickers (Rosenberg, 2003).

The walking time to carry harvested crops constitutes a significant non-productive part of the harvesting cycle for several fresh-market crops, like strawberries (Figure 1), raspberries, blackberries, and table grapes. For strawberries, walking time has been measured to reach up to 22% of the total harvest time (Khosro Anjom et al., 2018); higher inefficiencies are often reported, anecdotally.

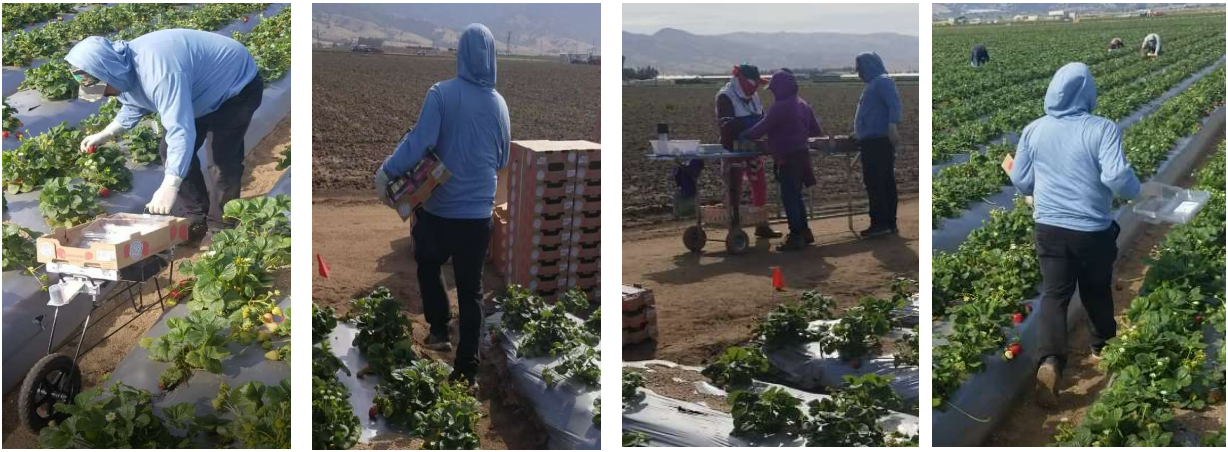

*Figure 1. The working cycle of manual harvesting in the open commercial field: a) the picker is picking strawberries inside the furrow; b) the picker transports the full trays to the collection station on the headland; c) the picker loads the trays in the collection station; d) the picker takes an empty tray back to resume picking.*

In this work, a collaborative robotic system (aka, co-robotic system) was investigated to assist in such harvesting operations by transporting trays, with strawberries as a case study (USDA REEIS, 2013). During the proposed robot-aided harvesting, each picker walks inside a furrow, harvests ripe fruits, and puts them in a standard-sized tray located on a special instrumented cart (Figure 2.a), in the same way as in all-manual harvesting. These carts are equipped with load cell sensors to measure the weight of the tray and a GNSS (Global Navigation Satellite System) module to record the geodetic locations of the carts (Khosro Anjom et al., 2018). The cart sends data wirelessly in real-time to a computer in the field (we refer to it as the "operation server"). Software running on the server predicts when and where a tray will become full (Khosro Anjom & Vougioukas, 2019). A full tray results in a tray-transport request to the scheduling software running on the server, which dispatches a team of crop-transport robots to serve those requests. The robots travel between the collection station and pickers to bring empty trays (Figure 2.b). The picker walks a small distance to the robot, loads the full tray, gets an empty tray, and pushes a button to command the robot to travel back to the collection station (Figure 2.c).

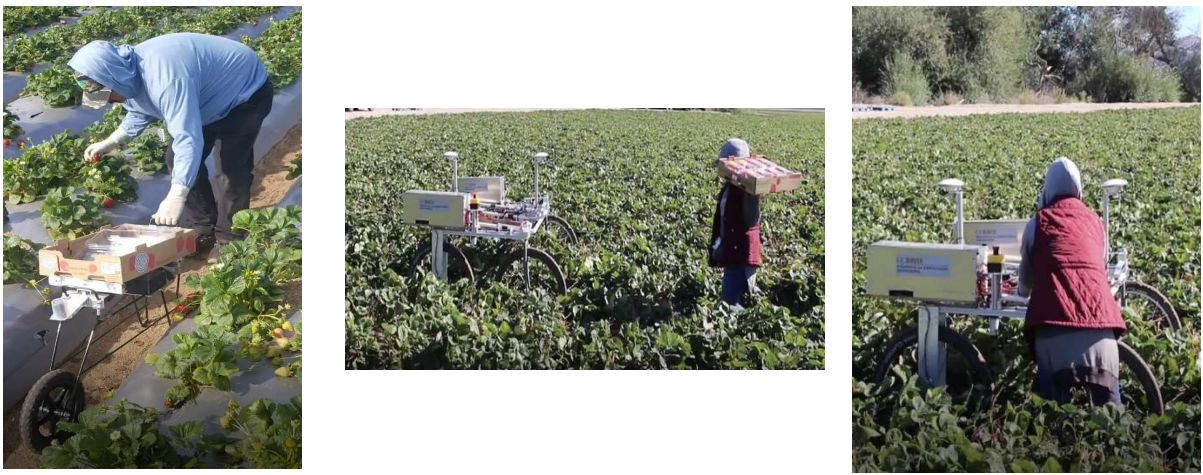

*Figure 2. The working cycle of co-robotic harvesting in the open commercial field: a) the picker is picking strawberries inside the furrow in the same way as in all-manual harvesting; b) the picker walks a small distance to the serving robot; c) the picker loads the trays on the robot;*

Given the large sizes of commercial harvesting crews (e.g., strawberry harvesting in California involves crews of twenty to forty people) and the expected cost and complexity of deploying equally large numbers of robots, this work explored an operational scenario in which a crew of pickers is served by a smaller team of robots. Thus, the robots are a shared resource with each robot serving multiple pickers. Given that robots travel at relatively low speeds, for safety purposes (in our case, $0.5 - 1.5$ m/s), and that the distance to a picker can be up to 100 m long, if the robots are not properly scheduled, robot sharing among the workers may introduce non-productive waiting delays between the time when a tray becomes full, and a robot arrives to collect it. Hence, *efficient robot dynamic scheduling* is essential to ensure that the overall reduction in walking time is larger than the waiting time introduced by robot operation.

Two main types of scheduling exist: reactive and predictive. In reactive scheduling (Blazewicz et al., 2019), a machine/vehicle/robot at the collection station is allocated to a task only after the scheduler receives the task request. In the context of tray-transport robots for harvesting, reactive scheduling refers to the situation where a robot starts traveling to a picker when the picker's tray becomes full. Seyyedhasani et al. (2020b) showed that when tray-transport robots were scheduled reactively, picker waiting time was reduced when the robot-to-pickers ratio increased. However, above a certain ratio, adding more robots did not reduce further the waiting time. The reason is that a picker's waiting time is at least as much as the time needed for a robot to travel the distance from the collection station to the picker.

Predictive scheduling policies incorporate information about future demand into the scheduling (Ritzinger et al., 2016). In the context of harvesting, 'future demand' refers to knowledge about when and where a worker's currently used tray will become full, giving rise to a tray-transport request. If the time and location are known in advance, a robot can be dispatched – and start moving toward that location - before the tray becomes full; hence, waiting times due to robot travel can be reduced or eliminated.

Such an approach was developed and tested in simulation by Peng & Vougioukas (2020). A robot-aided strawberry harvesting simulator was developed and a predictive scheduler was presented that knew exactly the times and locations of the tray-transport requests, and used exact and heuristic methods to compute the optimal schedule. The results showed that when the ratio of robots to pickers was high enough, predictive scheduling increased the harvesting efficiency of all-manual harvesting much more than reactive scheduling did (24% vs. 15%).

However, in reality, the locations and times of tray-transport requests contain uncertainty because of stochastic picker work-rate and varying – unknown - yield density (Khosro Anjom & Vougioukas, 2019). Uncertainty can be detrimental for predictive scheduling algorithms that assume perfect information (Bertsimas & Ryzin, 2017; Blazewicz et al., 2019). Hence, in this paper, a dynamic stochastic scheduling algorithm is used to account for the prediction uncertainty and improve the performance (Bent & Van Hentenryck, 2004; Blazewicz et al., 2019; Ichoua et al., 2006).

The goal of this paper is to present the development of an integrated harvest-aiding co-robotic system – with emphasis on online stochastic predictive scheduling - and the results from its evaluation in real-world commercial strawberry harvesting. The contributions of the paper are summarized as follows:

(1) Software and hardware development of an integrated co-robotic harvesting system that includes instrumented picking carts, autonomous crop-transport robots, a field computer running scheduling and operations management software, and a communication system that connects all the above.

(2) Adaptation of a stochastic scheduling algorithm for online crop-transport robot scheduling based on the prediction of transport requests, with application in strawberry harvesting.

(3) Field experiments during commercial strawberry harvesting and assessment of system performance.

The rest of the paper is organized as follows. Section 2 surveys the related work of the in-field logistics operations. In section 3, the physical implementation of the co-robotic system is presented. Section 4 introduces the mathematical modeling of predictive stochastic scheduling for the robot team in the context of co-robotic harvesting and how the scheduling problem is solved in an online fashion. Section 5 presents the experimental design and the analysis of experimental results from simulating and real-world commercial strawberry harvesting using the developed system. Finally, section **Error! Reference source not found.**, presents the main conclusions of this work and suggests directions for future research.

## 2. Related work

The use of mobile robots to carry the trays that pickers use, and the related robot scheduling problem, bear similarity to the use – and scheduling - of Autonomously Guided Vehicles (AGVs) in flexible manufacturing systems (FMS), and was inspired by it (USDA-REEIS, 2013).  A typical FMS consists of work machines that feed on materials and parts and produce product components; a material handling system (MHS) which moves materials and parts, and a central control computer which orchestrates material movements and production flow (Buzacott & Yao., 1986). A modern MHS will utilize AGVs to perform material transport operations, such as carrying materials to the machines' input buffers, removing parts from the machine output buffers and transporting them to other machines (for further processing), and delivering finished parts from output buffers to a collection station. Typically, the AGVs are scheduled and dispatched based on the product dispatch times of the work machines, which are stochastic, due to reasons such as machine breakdowns, stochastic processing times and unexpected releases of high priority jobs (Ghaleb et al., 2020). Existing approaches used to solve the FMS scheduling problem includes genetic algorithms, stochastic dynamic programming, integer programming, heuristic algorithm and so on (Yadav & Jayswal, 2018). Similarly, in robot-aided harvesting, the unknown and spatially variable crop distribution and the highly stochastic harvesting speeds of the human pickers result in stochastic "processing time", i.e., the time it takes a picker to fill a tray (which must be transported) is stochastic. However, our problem is different, because the workers are not stationary – like work machines are - so the transport requests are stochastic both temporally, as well as spatially. Additionally, workers can transport their own trays, and therefore, the transport robots can afford to reject transport requests, something that AGVs cannot do, in an FMS setting. Hence, the scheduling paradigms for AGVs operating in manufacturing environments cannot be applied directly in our application.

The crop-transport robot operation also bears resemblances to automated warehouse logistics systems, in which humans and robot work together. Given an order for a certain product, a mobile robot is dispatched to the location of the rack that contains the product, it lifts the rack and moves it to a station where human workers pick items and place them in boxes (Weidinger et al., 2018). The robots relieve workers from walking unproductive long distances in the warehouse and increase the order processing productivity (Olsen & Tomlin, 2020). In such warehouses, many orders may arrive over short time intervals, driven by an online ordering system (Li et al., 2020). The robot scheduler must handle the dynamic product demand, while considering the stochastic and time-varying work speed of human operators, which affects the time when a robot must return to a worker to retrieve the rack and move it to its original location (Pasparakis et al., 2021; Wang et al., 2021). Wang et al., (2021) developed an online prediction system that predicts the workers' picking rates using stochastic models and schedules the mobile robots to serve them with stochastic dynamic programming; the algorithm balances picker workloads and improves the overall productivity. In automated warehouses, just like in robot-aided harvesting, the transport requests are dynamic and robot scheduling is coupled to human working speeds. However, in contrast to our problem, the dispatch destinations of the mobile robots are known/deterministic and stationary; only the return times to the workers are stochastic. In our application, both the locations of the robot destinations and the required times of arrival are stochastic, because they are coupled spatiotemporally to the pickers' stochastic harvesting activities.

To handle the spatiotemporal uncertainty of transport requests, we applied a fast and suboptimal scenario sampling-based algorithm, Multiple Scenario Approach (MSA), developed for the partially dynamic vehicle routing problem with time windows (R. W. Bent & Van Hentenryck, 2004). In the planning stage, MSA samples the uncertainties and generates multiple scenarios for the tray-transport requests. Given each sampled scenario, deterministic scheduling was solved quickly and computed a near-optimal schedule for the robots. At the decision instant, the actual schedule that is followed by the robots is computed from the sample scenarios using a consensus function (R. Bent & Van Hentenryck, 2004 June). This schedule is the most consistent with the optimal schedules of all the sampled scenarios. This methodology can be adapted and applied to different online stochastic combinatorial optimization problems by building two case-dependent modules: a scenario sampling function to generate multiple deterministic scheduling scenarios, and a scheduling solver for each sampled deterministic scenario (Pillac et al., 2013).

## 3. Design of the co-robotic harvesting system

Strawberries are planted in rows with furrows between them that accommodate human and machine traffic (**Error! Reference source not found.**.a). The field headlands are used for collection/parking/inspection stations and traffic of people, forklifts and trucks involved in the handling and transportation of the harvested crops. A typical harvesting block consists of approximate 80~120 rows, of about 100 meters in length. Before harvesting, collection stations and empty trays are placed at one side of the field, at the headland. The crew (say, *N* people) start picking as a team in the first *N* rows on the left or right side of the block. Although there may be several collection stations, the one that is closest to the crew is the active one. To reduce the walking needed to transport full trays to the collection station, the standard harvesting practice is to have workers start picking from the middle of the field block and walk outward, toward one of its edges. Once that section of the block is harvested, the collection stations are moved to the other edge, and the other section (half of the block) is harvested. This method essentially 'splits' a harvesting block into two sections (an example is given in **Error! Reference source not found.**.b, where there is an upper and a lower block, above and below the blue dotted line, respectively). The field can be modeled using points at the edges of each furrow; two points to represent the middle line, and points for each collection station.

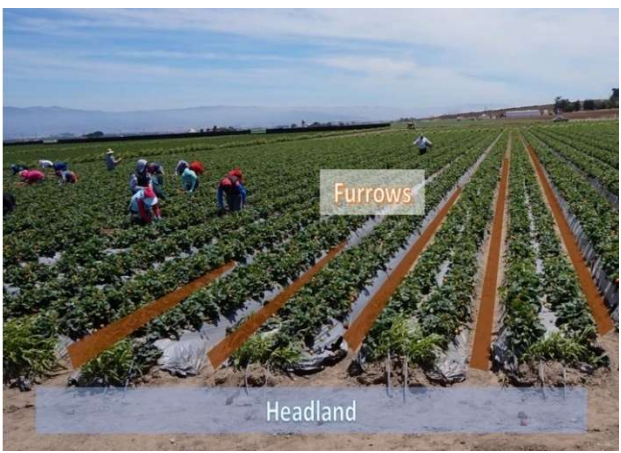 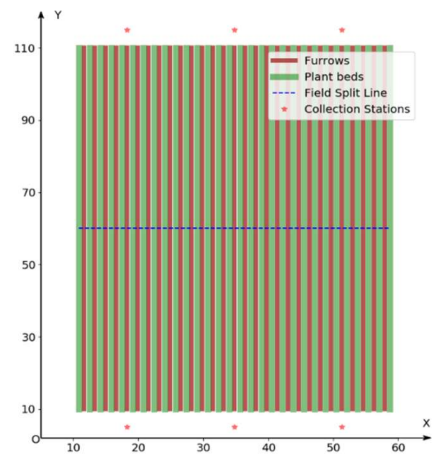

*Figure 3 - a) Layout of a typical raised-bed strawberry field; 3b) schematic figure of the strawberry harvesting field block with two sections (upper and lower); furrows; plant beds; field split line, and collection stations (Peng & Vougioukas, 2020).*

In the envisioned crop-transport robotic aided harvesting, each picker enters an unoccupied furrow to start picking strawberries selectively from the plants on the raised beds on each side of that furrow. When a certain fill ratio (percent of tray filled - FR) is reached (Peng & Vougioukas, 2020), the picker will press a request push-button on their instrumented cart. The button allows a picker to decide for themself if they want a robot to carry their tray. For example, if a picker prefers to walk to deliver a tray - to take a break from stooped work - they can do so. The button also helps to establish a simple communication protocol between the workers and the robotic system: a transport request is initiated and is either accepted by the system or rejected. The automated weighing system is used by the transport-request prediction module, after the button has been pressed. The scheduling system signals the picker (LEDs on their instrumented carts) if their transporting request will be served by a dispatched robot or rejected by the system. If the picker will be served, the dispatched robot starts from the active collection station, drives with an empty tray to the assigned picker's full tray location, waits for the picker to switch empty and full trays, and takes the full tray back to the active collection station, where it waits for the next dispatching. If a tray-transport request is rejected, the picker transports the full tray to the collection station, just like in manual harvesting.

The co-robotic harvest-aiding system comprises three sub-systems: instrumented carts, robots (aka FRAIL-Bots: **F**ragile c**R**op h**A**rvest-a**I**ding mobi**L**e ro**bots** - USDA REEIS 2013), and an operation server. The carts used by pickers weigh approximately 2.2 kg and our instrumented carts should have the same form factor and not be significantly heavier, to be accepted by pickers (4 kg max). The volume and weight constraints introduced a battery size – and available energy – constraint. Also, small amounts of data from each cart must be transmitted to the field computer/operation server over distances that span typical fields (> 300

m) at rates of approximately 1 Hz; communication must be bidirectional, since the scheduler must inform pickers if their requests are rejected or will be served. The robots must also communicate in real-time with the operation server to send their state and receive dispatching commands and reference paths. Robot collision avoidance on the headlands is performed centrally on the server and thus the robot-server communication system must have high-bandwidth. The complexity of the software running on the carts, robots and operation server requires a distributed software architecture that can provide real-time performance.

The system architecture is shown in Figure 4, and is described next in greater detail.

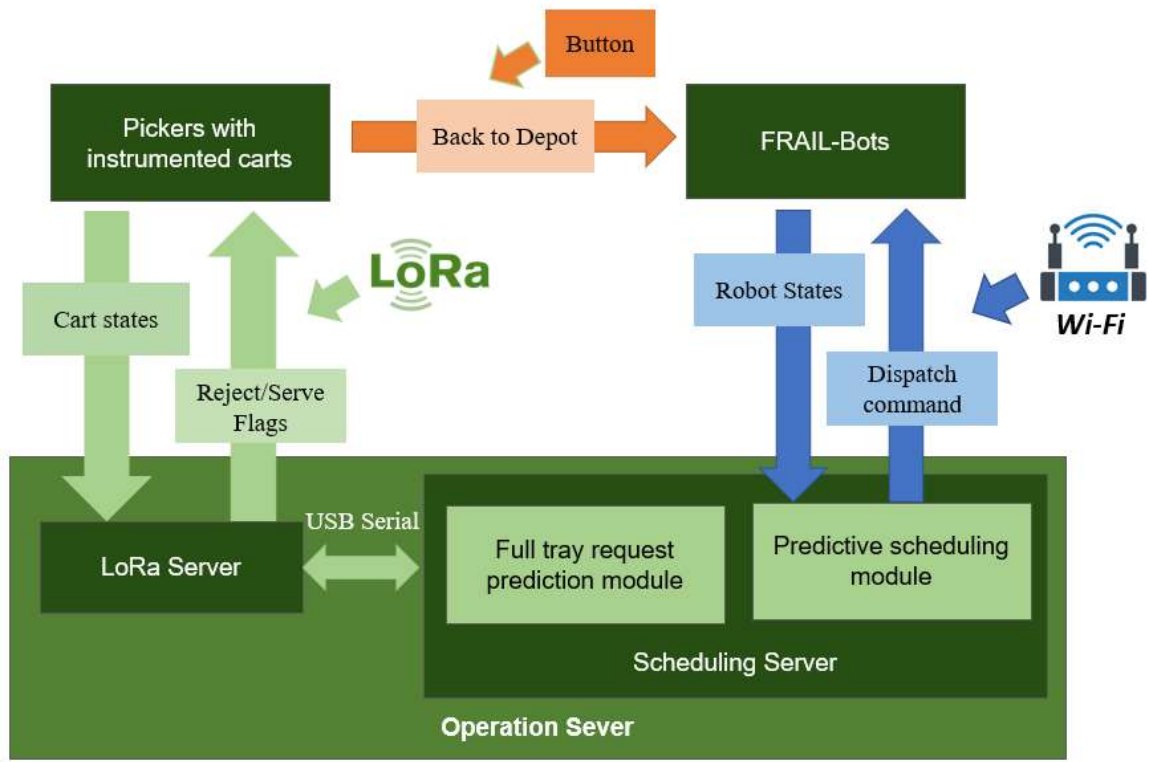

*Figure 4. Diagrams of harvest-aiding system components and their communications*

The data from instrumented carts are transmitted wirelessly to the server module that runs on a field computer at a collection station through LoRa ("LoRa", 2020). The LoRa server, scheduling server, and FRAIL-Bots communicate with each other using a ROS network (Figure

4) that utilizes different physical layers. The LoRa server module is connected through a USB cable to the scheduling server computer that works under the same Wi-Fi network with FRAIL-Bots Each functional module in the ROS network is implemented and packaged into a ROS node which can subscribe and advertise the ROS messages from the other nodes. On the scheduling server, the tray request prediction module receives data from the cart and generates predictions of full tray requests. Given the subscribed ROS messages of robots' states and the predicted full tray requests, the predictive scheduling module calculates an optimized schedule. Then, the scheduling module publishes the dispatching commands, which are received by the available robots. After the robots arrive at the predicted full tray location, the picker will load their harvested full trays and take the empty tray from the arrived robot. Then they press a button on the robot to signal it back to the collection station. When the robots arrive back at the station, they will wait for the worker at the station to unload the full tray and replace it with an empty tray.

## 3.1.    Sub-system I: instrumented picking carts

The instrumented picking cart was modified and fabricated from a standard strawberry harvest cart (Figure 5). During harvesting, the weight of the strawberries inside the tray located on the cart is measured using two load cells underneath the supporting frame. The cart's geographical coordinates are received from a GNSS receiver module (Piksi-Multi, Swift navigation, US) that incorporates WAAS (Wide Area Augmentation System) corrections in real-time. An IMU (BM160, Bosch, German) is integrated on the Piksi-Multi to measure the instantaneous motion of the cart which is used to filter the weight measurements. The weight and GPS locations are transmitted to the scheduling server by a wireless LoRa module (RFM96W LoRa Radio, Adafruit, US) located on the control board. The LoRa is a low-power wide-area network protocol, which uses license-free sub-gigahertz radio frequency bands ("LoRa", 2020). A momentary contact button is available for the picker to notify the system that she/he wants to be served by the robot.

The cart messages, composed of measured weight, GPS locations, and the button state, are transmitted at 1 Hz interval to the LoRa module on the operation server. An SD card module installed on the control board is used to store all the sensor data during harvesting. Two LEDs (red and yellow) are used as indicators to communicate informative signals to the picker: a yellow LED turns on when the tray transport request has been assigned to a robot; the red LED turns on if the request is rejected, and the picker needs to transport the tray by themself.

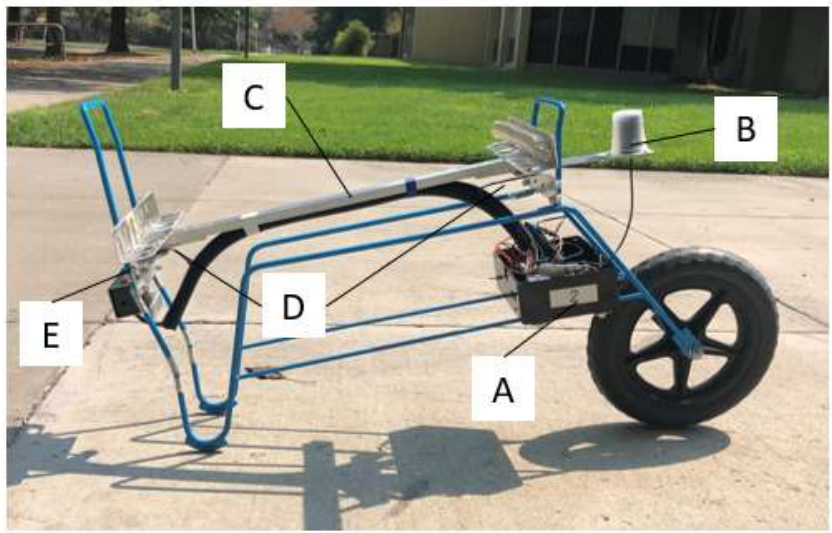

*Figure 5. The instrumented picking cart: A. Control box with Arduino Due, LoRa module, battery, and SD card logger inside, Piksi Multi GPS unit; B. GPS antenna; C. Supporting frame on the top of load cells; D. Load cells; E. Momentary push button, yellow and red LEDs.*

### 3.2. Sub-system II: FRAIL-Bot

Two identical crop-transport robots (aka FRAIL-Bots) were designed and built for this work (Figure 6). The bill of materials for each robot is approximately 10,000 USD; fabrication cost is not included. Constraints related to budget, available time and field deployment restricted the number of robots to two. Still, the stochastic scheduling approach and the entire system is applicable – and can be tested – with two robots, and reasonable robot-picker ratios can be achieved by using a crew size of six to eight people. The robots are designed to straddle the bed and occupy two furrows when driving inside the field. To avoid any interference of the robot with pickers in adjacent rows, pickers need to be spaced two furrows apart (with one empty furrow between them). This arrangement was acceptable by the growers and the pickers (and was actually used anyway during the pandemic, for distancing purposes).

Each robot works under supervised autonomy and its collaborative operation is governed by a finite state machine (Peng & Vougioukas, 2020). The hardware components for the FRAIL-Bot are labeled in Figure 6. The robot weighs approximately 50 kg, and it is driven by two DC motors with gearboxes and incremental encoders attached to the rear wheels (D and E). The steering system is integrated with two screw drives and angle sensors attached on their rotation axis (F and H). Two GPS module antennas (Swift navigation, US) are installed for getting the position and heading of the robot in open fields (C). An emergency stop button (I) is installed on the side of the robot to stop the driving system. A return button (E) on the front of the robot is used by the pickers to signal the robot that the full tray has been loaded and the robot must drive back to the collection station. The electronic devices including batteries, mini-computer (Intel NUC, Intel Inc, US), driving motor controllers for rear-wheel motors, steering motor controllers, and two GPS modules are installed inside two wooden boxes (A and B).

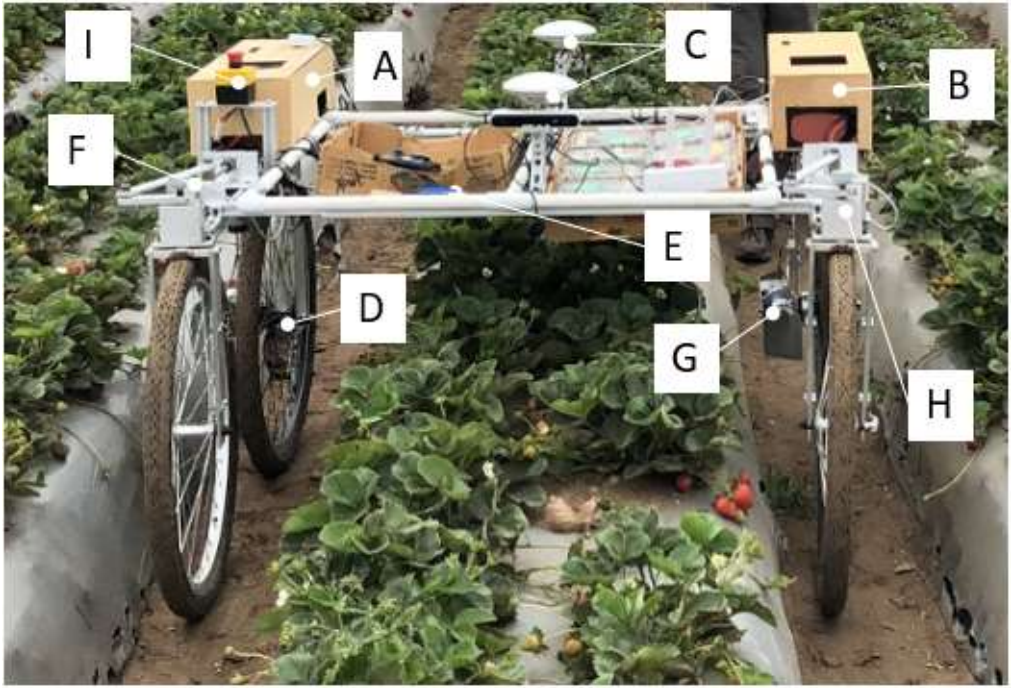

*Figure 6. Components of FRAIL-Bot: A. control box-I with a mini-computer, battery-I, motor controllers for the two rear driving motors and two steering motors B; Control box-II with two GPS modules; C. GPS antennas; D, G. DC motors with gearbox and incremental encoders; E. Return button; F, H. Steer-driving system; I. Emergency button.*

The software architecture on each assigned FRAIL-Bot is shown in Figure 7. The FSM node first subscribes to the schedule message from the operation server including the dispatching time and dispatching location, from the operation server (explained in the section 3.2). When the dispatching time is reached, the navigation node generates the planning path from the current location of the robot to the assigned location inside the row. Given the planned path and the current robot location and heading, the path tracking module continuously outputs the control command to the motion control node, which converts the motion command of the robots to the commands of driving motors. A robot localization node fuses the subscribed sensor information, including GPS measurements, IMU data, and wheel odometry to estimate the pose and publishes the current pose of the robot into the ROS network.

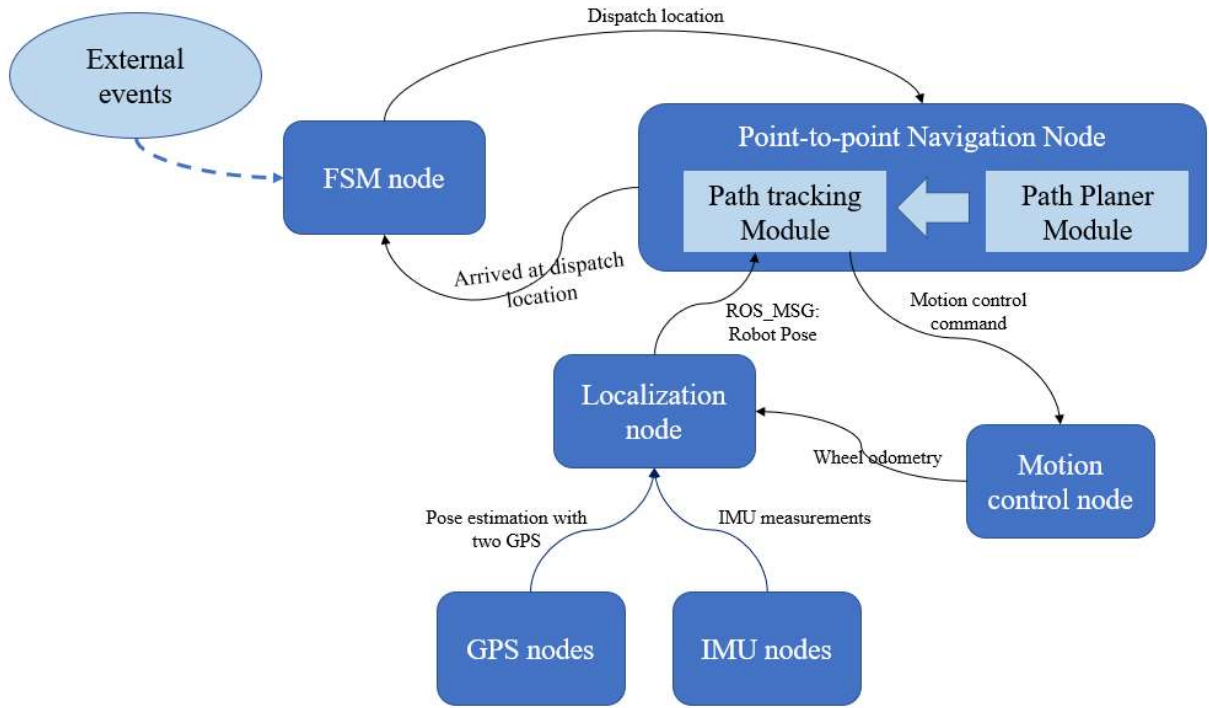

*Figure 7. The architecture of the FRAIL-Bot software under ROS*

The activity of FRAIL-Bot is guided by a finite state machine (FSM) introduced in our previous work (Peng & Vougioukas, 2020). The FSM implemented in the ROS node is shown in Figure 8. The state transitions are based on current states and designed external events (green oval in Figure 8) during the harvesting activity. Each FRAIL-Bot navigates autonomously from the collection station to the picker it will serve after receiving the dispatching command from the predictive scheduler. Upon arrival at the dispatch location, the robot waits until the picker fills their tray (if it is not full upon arrival), loads the full tray, takes an empty tray from the robot, and presses the momentary contact button to command the robot back. The robot navigates back to the collection station and enters the state "IDLE_IN_QUEUE", i.e., it waits for a worker at the collection station to remove the full tray and place an empty tray on the robot. Afterwards, this worker presses the momentary contact button to have the robot transit to the "AVAILABLE" state, where it waits to be dispatched again.

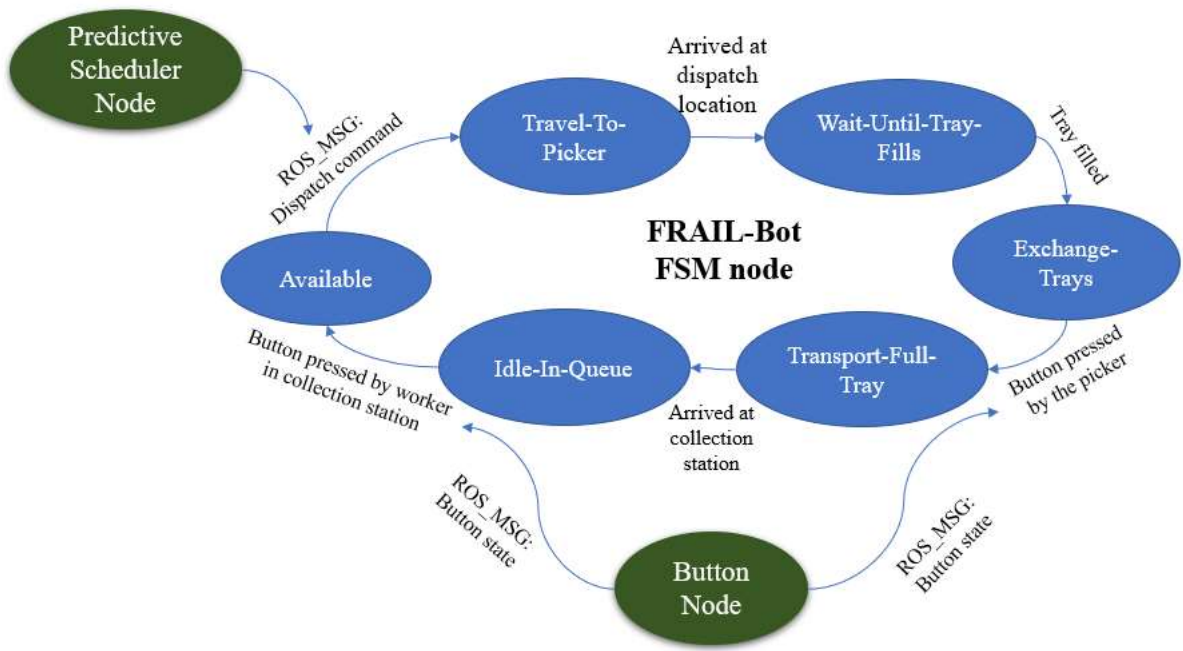

*Figure 8. FSM of FRAIL-Bot in the harvest-aiding system*

### 3.2.1. Motion control

The hardware diagram for the motion control node of the FRAIL-Bot is shown in Figure 9**Error! Reference source not found.**. A dual-channel motor controller is used to drive the two rear motors by UART serial communication. The left and right steering systems are driven by two DC motor control boards (1065B, Phidget Inc, Canada) with the PID controllers based on the feedback from angle encoders.

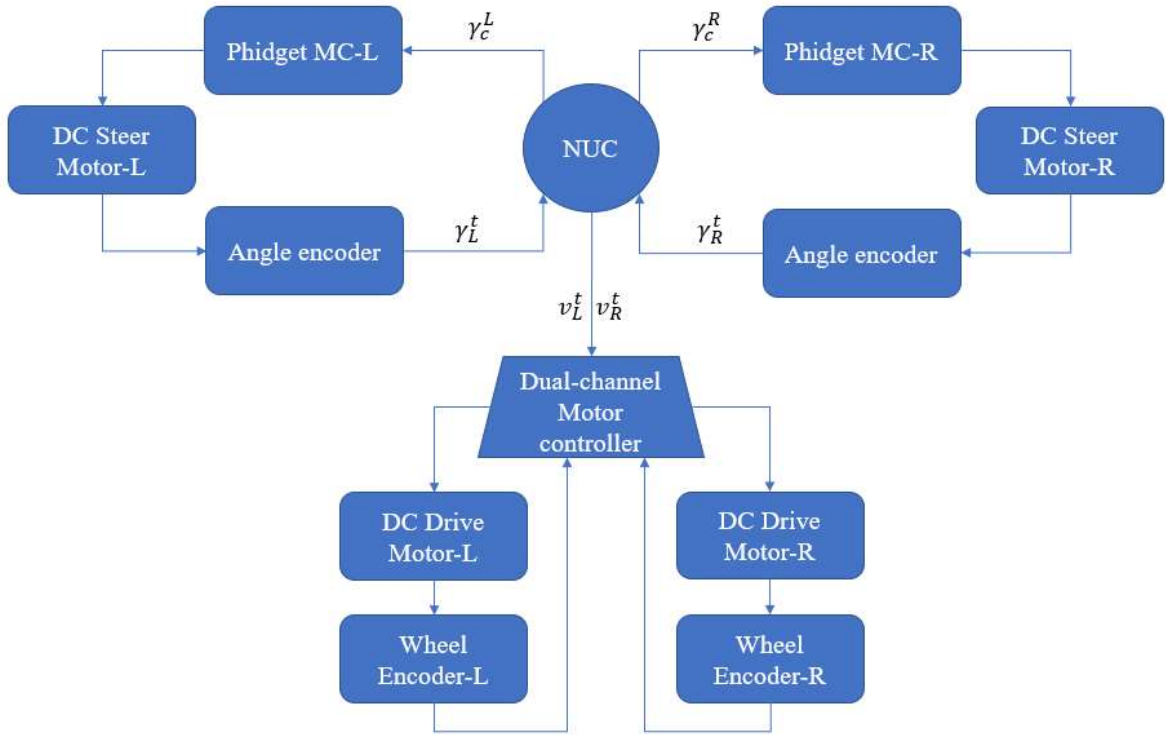

*Figure 9. Hardware diagram for FRAIL-Bot motion control*

The Ackerman model (Figure 10) was applied for the motion control of the FRAIL-Bot to have it maneuver smoothly (minimum skidding) in the field navigation. Given the motion command, linear velocity ($v_c$) and steer angle ($\gamma_c$), the steer angle command ($\gamma_c^L, \gamma_c^R$) for left and right steering systems can be calculated using the Ackerman model.

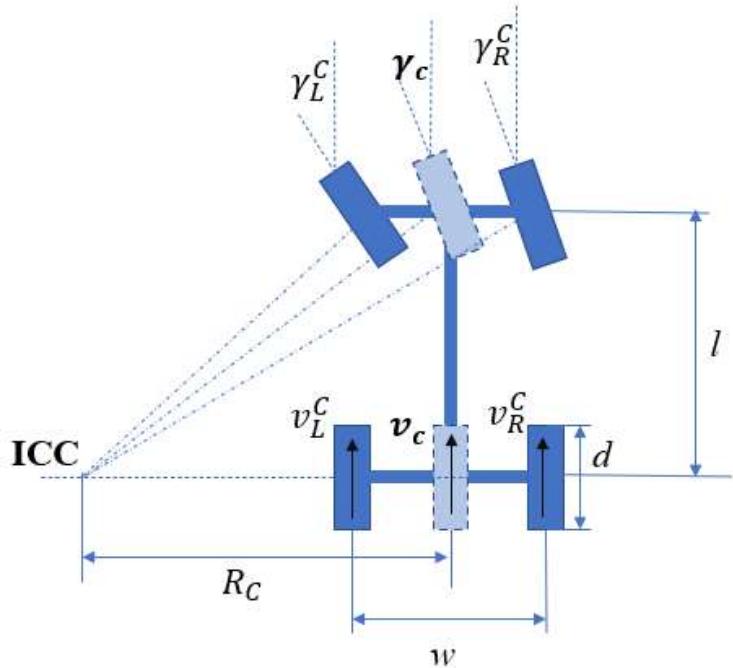

*Figure 10. Ackerman model for the calculated control command of FRAIL-Bot*

The response time of the steering angle control system is slower than the response of the speed control system of the wheel driving motors. Therefore, the wheel motor speeds are commanded to "follow" the robot rotational velocity that corresponds to the sensed steering angles of the front wheels (cascade control). The real-time control commands for steering motors are calculated based on the motion control command ($v_c$, $\gamma_c$). The rear motor commands ($v_c^L$, $v_c^R$) are calculated from the current steering angle ($\gamma_c^t$) and the linear velocity command ($\boldsymbol{v_c}$).

### 3.2.2. Localization

The localization node of the FRAIL-Bot is implemented with the EFK node on the ROS package, named "robot-localization" (Moore & Stouch, 2016) by fusing the measurements of two GNSS modules, wheel odometry, and IMU as shown in Figure 11. The sensor information, including 2D pose estimation from the GNSS receivers, wheel odometry on the robot frame

(base link frame), acceleration, and angular velocity on the IMU frame are packaged and published as ROS messages.

The RTK (Real Time Kinematic) solutions of two GNSS modules build a 2D pose estimation of the FRAIL-Bot. The GPS antennas on FRAIL-Bots are above the strawberry plants and the strawberry field is normally in an open area without nearby high buildings or tree canopies which might attenuate or occlude satellite radio signals. Thus, the RTK solutions are stable for most of the time. The Internet-based (NTrip) RTK system was used to get the geodetic solution of the rear GPS antenna under the transmission protocol of Radio Technical Commission for Maritime (RTCM) (Weber et al., 2005). The NTrip casters used in California are managed by UNAVCO community service (Bendick, 2012). As the GPS location of the NTrip caster is well surveyed on the geodetic frames before releasing, the RTK rover solutions corrected by the same base station in a mapped field are consistent over time. The front GPS also works in RTK mode with the real-time corrections from the rear GPS that works as the moving base station based on the same satellites' observation over their antenna attitudes (Swift Navigation, 2020). Using this feature, the attitude of the robot, heading angle, in the geodetic frame from the front GPS solution was obtained. Combining these two results, the complete localization solution of the robot in 2D space can be obtained in a mapped field.

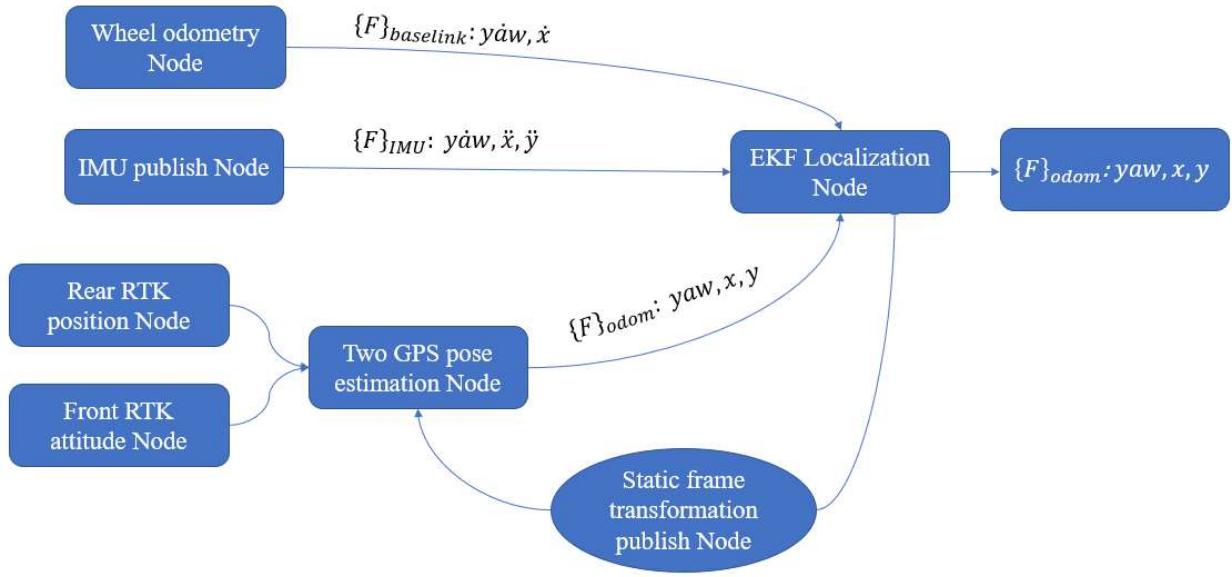

*Figure 11. FRAIL-Bot localization with robot-localization EKF node*

### 3.2.3.   Field navigation

To transport trays during harvesting, the FRAIL-Bots travel back and forth from fixed "parking" locations next to the collection station on the headland to dynamic locations inside the field. A field map (Figure 3.b) with the geometric information of the plant beds and the collection station is essential for the FRAIL-Bots to generate feasible paths to their serving locations. The map is built with the RTK system and includes the end of bed points in the vicinity of the headland and pre-allocated collection stations.

To implement point-to-point navigation in the field, the path planner needs to generate a smooth path that allows the robot to enter the row with heading parallel to the row. A speed profile is also generated to vary the robot speed at different sections of the path. When driving on the headland the robot often needs to execute maneuvers of large curvature and thus must travel at low speed. Also, when the robot is near a picker, the robot needs to run at a low speed, for safety purposes. Inside the rows, the robot can run at a higher speed. The path is planned as soon as the robot receives the dispatching command (Figure 12).

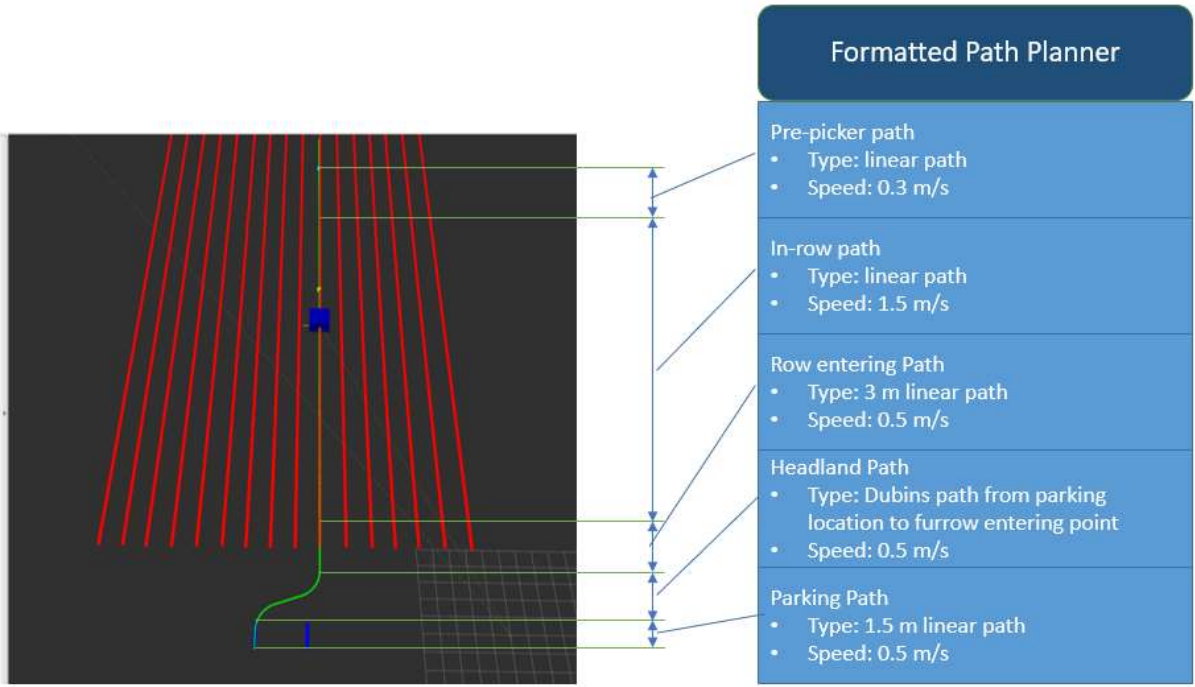

*Figure 12. An example of a formatted planning path from the robot parking location at the collection center to the dispatching location inside the row*

A pure-pursuit algorithm was used to have the robot track the path. The algorithm parameter, look ahead distance, is well-tuned and set differently given the designed speed profile. The look-ahead distance was tuned given different planned speed along the trajectory.

## 3.3. Sub-system III: operation server

The hardware of the operation server sub-system is composed of two parts: a LoRa server board and a scheduling server computer (Figure 13). The LoRa server board is connected to the server computer by a USB cable. It collects the cart states from the distributed instrumented carts in the field and publishes the received states as ROS messages. The FRAIL-Bots publish their states in the ROS network through a local Wi-Fi. The scheduling server module, running on the server computer, integrates the cart states and robot states to formulate and publish an online schedule message in the ROS network. FRAIL-Bots directly subscribe to the ROS schedule messages and execute the dispatching decisions. The LoRa server module also subscribes to the ROS schedule messages and transmits them to each instrumented cart through LoRa.

### 3.3.1. LoRa communications module

The LoRa module on the cart and the server had adjustable parameters for bandwidth, spreading factor, code rating, and transmission power, which needed to be adjusted (Cattani et al., 2017). Based on field testing, the following parameters met our requirements: bandwidth at 250 kHz, spreading factor at 6, code rate 4, and transmission power at 14 dBm.

During transmission, each module can only work in a single channel, which means that the server can only get the message from one cart even if multiple carts send their messages at the same time. Our goal was to have the communication channel of the server evenly shared by all the carts in the time space. Based on our testing, the transmission of each cart state to the server takes around 100ms, so given $N$ carts, the server takes at least $N*100$ms to receive the cart states from all the carts once. If the size of the picking crew is over 10 (a typical number is 20 to 25), a LoRa gateway with multiple channels would be needed to meet our transmission requirements. In our experiment, we deployed two robots and a picker crew with 6 pickers, so a single channel LoRa server was used. A centralized time-split network communication protocol was developed in our work, as shown in Figure 13. Each period, the server broadcasts a reporting 'command' which is a short message containing the ID of a cart and a 'serve' or 'reject' flag updated from the predictive scheduler on the server computer. After that, the LoRa server waits for 100ms for the polled cart to report their cart states. The cart with that ID broadcasts its message after receiving the command message. The other carts will ignore the messages on the channel and keep waiting for their ID to be requested to report their states. If the server received the message from that cart or the wait time expires, it will move on to the next cart.

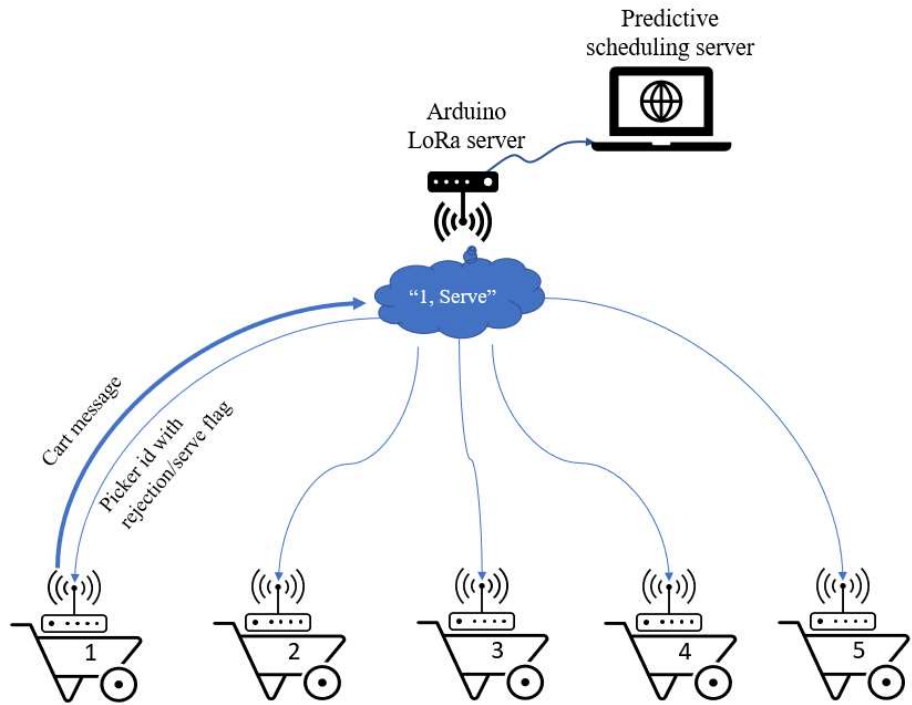

*Figure 13. The centralized topology of communication between LoRa modules on each cart and the server*

The above-described functionality of the LoRa server module is packaged into two ROS nodes (Figure 14):

a) Cart-states-pub node: It receives the data of cart states from the LoRa module on the server board, packages them into ROS messages, and advertises them to the server computer through a USB cable.

b) Server-reject-sub node: It subscribes to the 'serve' or 'reject' flags advertised by the predictive scheduling node and transmitted to each cart via LoRa.

### 3.3.2. Scheduling server module

Each functionality of the scheduling server module running on the server computer is packaged into a single ROS node. This module plays a centralized role in the whole system shown in Figure 14. The nodes running on the server computer are briefly explained as follows.

a) Tray-request-prediction node: It subscribes to the cart messages from the LoRa server board and updates the prediction of picking parameters, harvesting rate, and moving speed while picking. The predictive requests are generated and published on the ROS network when a certain fill ratio of the tray is reached, and the request button is pressed by the pickers.

b) FRAIL-Bot-scheduling node: It subscribes to the robot states from the FSM node running on each FRAIL-Bot, and to the predictive tray transport requests from the tray-request-prediction node. Given these data, this node runs a stochastic predictive scheduling algorithm and advertises dispatching commands to the FRAIL-Bots, as well as the rejection and serve flags to the LoRa server boards. The online solver of this node is explained in section 4.

c) FRAIL-Bots coordination node: It functions as the traffic management for the robots in the shared area of the headland and is explained in section 3.3.2.2.

d) Operation visualization node: This node subscribes the messages from multiple nodes of different modules for visualization of the cart/robot states. It also provides some user interface to tune the parameters during the field operation, which is explained in section 3.3.2.1.

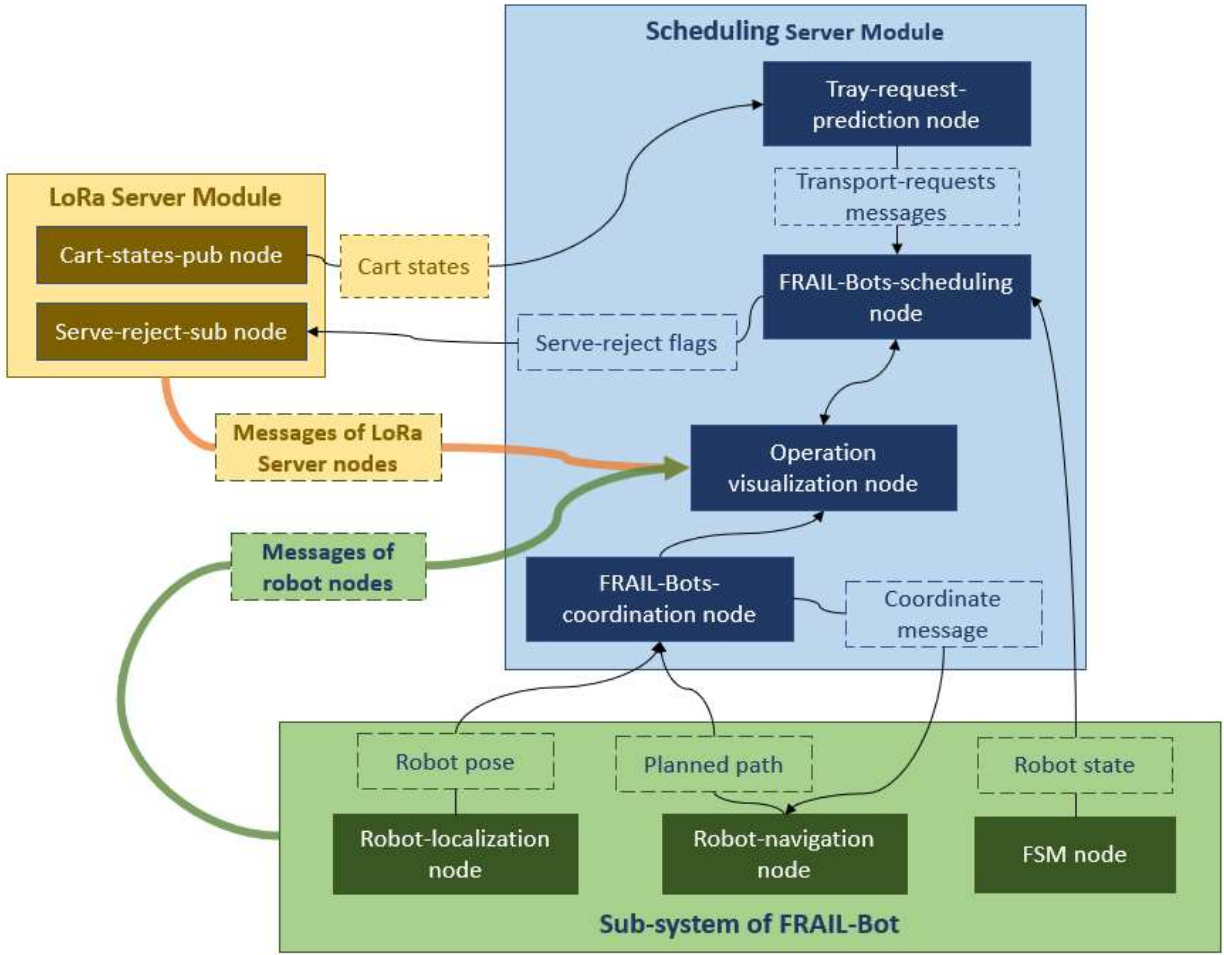

*Figure 14. The system architecture of the operation server modularized as ROS nodes.*

### 3.3.2.1. Operation visualization node – User interface

The user interface of the operation visualization module is shown in Figure 15. The sub-plot titled "Field Map" shows the current locations of the robots and the instrumented carts; the subplot "Measured Weights" displays the measured weight of each cart over time. The operational states of the carts are represented with four flags: "Full-tray", "Request", "Serving", "Reject". The "Full-tray" box turns green for 4 seconds after the picker lifts the full tray from the cart. The "Request" box color turns green if the robot-request button on that cart is pressed. The "Serving" box color changes to green when a robot is dispatched to that cart and "Rejection" is changed to green when the scheduling server rejects the transport request of that cart.

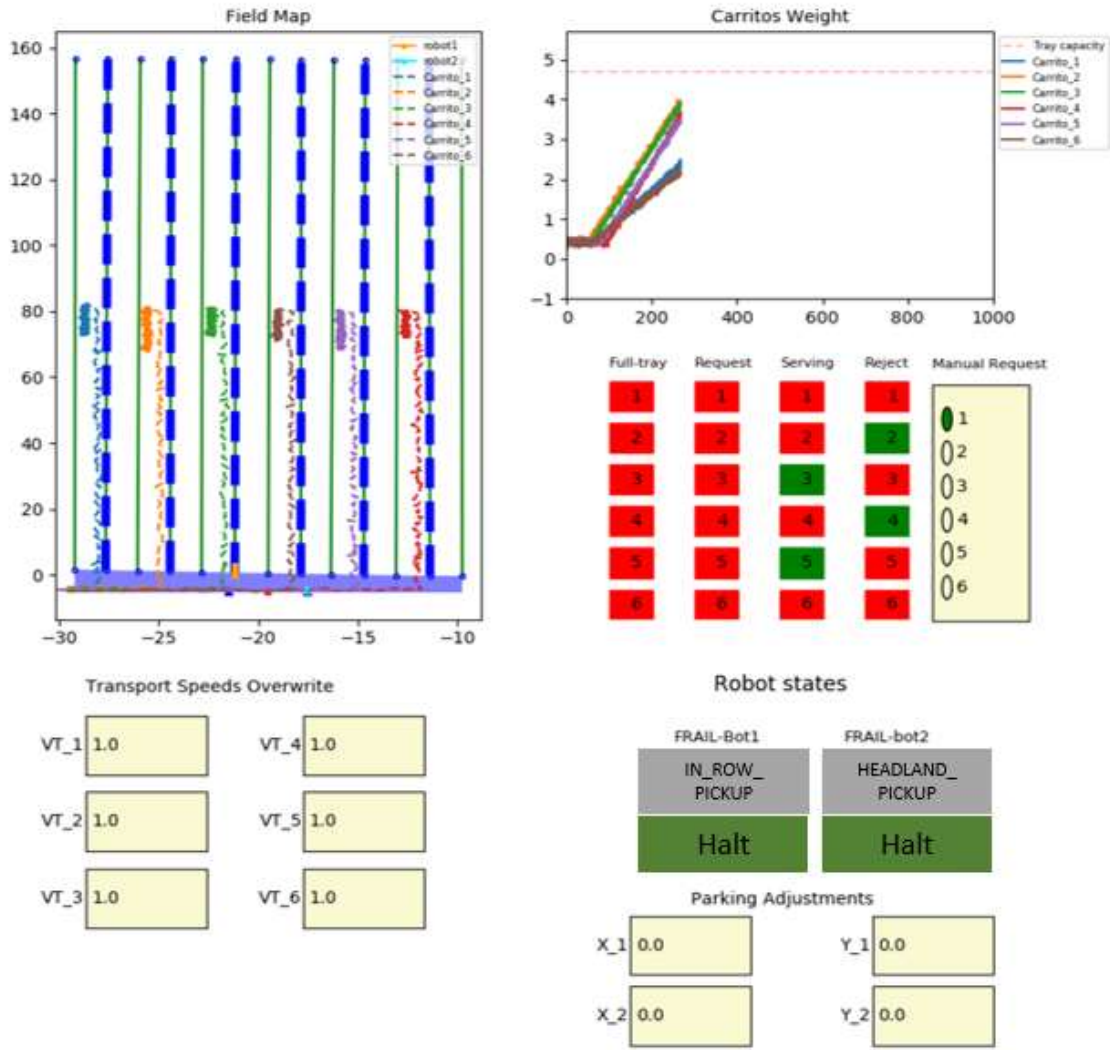

*Figure 15. System visualization for the states of instrumented carts and FRAIL-Bots and user interface for the parameters tuning.*

### 3.3.2.2. Robot coordination node

As the field headland space is shared by the robot team, a coordination node is required to prioritize the motion of one robot to solve the trajectory intersection. In this work, the harvest efficiency improvement was investigated after introducing the harvest-aiding system in the studied case. The function of the robot coordination node is mainly to coordinate the robots so that they do not collide with each other when their paths are projected to intersect on the shared headland space.

      The coordination node made the decision on which robot moves earlier based on the received planned trajectories and current locations of the two robots. When robots are driving on the headland, the robot coordinator node builds polygons encompassing the future paths from the robots' current pose to the end of their paths in the headland space, as shown in Figure 16. If the polygons intersect and one of the two robots has entered the intersected area, the coordinator will prioritize the robot inside the intersection area to go out of the intersecting area to avoid collision. As it is shown in the example in Figure 16, Robot-1 has entered the intersection area while Robot-2 has not yet. Given the planned paths and the robots' current poses, the coordinator predicts two instants for the robots: when they enter $(t_{r1}^1, t_{r2}^1)$ and exit $(t_{r1}^2, t_{r2}^2)$ the intersection area relative to current time instant. As Robot-1 (blue rectangle) has been inside the intersection area, $t_{r1}^1 = 0$ and $t_{r1}^2$ is predicted given the planned path. If the two robots' time intervals $[t_{r1}^1, t_{r1}^2]$ and $[max(0, t_{r2}^1 - margin), \ t_{r2}^2]$ intersect, the robot outside the intersection area (Robot2 in this case) will be commanded to stop before entering the intersection area, and wait until the robot time intervals do not intersect. A safety margin of 5 seconds is set when calculating the entering instant of the robot outside the intersection area.

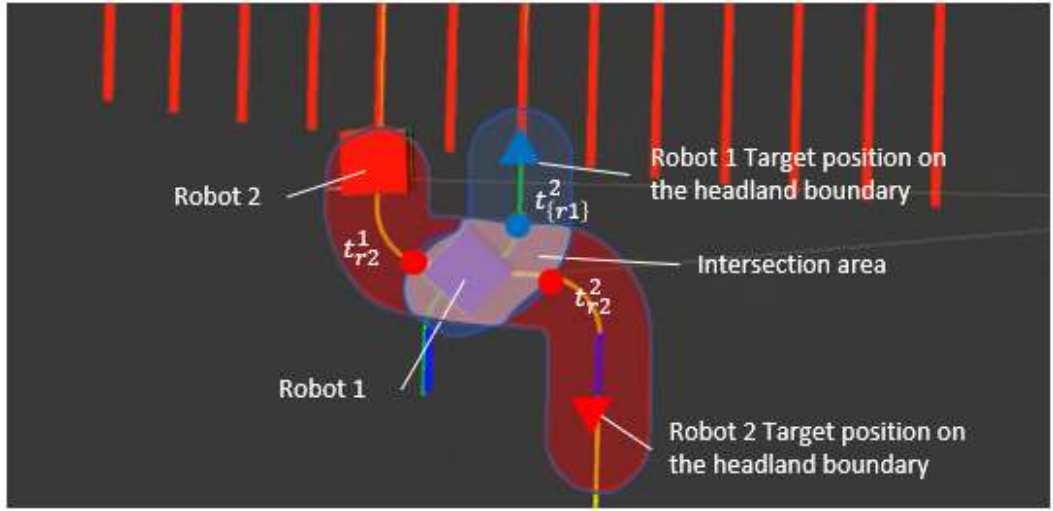

*Figure 16. The robot coordination node sweeps the robots' bounding polygons along their planned trajectories to predict possible collision; visualization is done using ROS-RVIZ*

## 4. Scheduling of FRAIL-bots under stochastic requests

In our previous work (Peng & Vougioukas, 2020), the predictive scheduling of crop-transport robots was modelled as a parallel machine scheduling problem (PMSP) with a release time constraint. The scheduler served all the requests, and its objective was to minimize the total waiting time of all requests.

In this work, two major changes were made to the scheduling problem: (1) the scheduler takes into consideration the inherent uncertainty in the predictions of the tray-transport requests, and (2) the scheduler is allowed to reject transport requests. The first change was necessary because the time needed to fill a tray and the corresponding distance travelled by the picker are random variables that follow stochastic distributions. The second change was also necessary, because, if all requests are served, a picker may wait for a robot to arrive, even if the waiting time is longer than the time it would have taken the picker to walk and deliver the tray themselves. Such a policy would be inefficient, and not acceptable by the pickers, who are paid based on the number of trays collected; long waiting times lead to fewer trays and lower pay.

The objective of the scheduler is to minimize the expected total non-productive time of all the transport requests. The formulation of the scheduling problem is given next.

In robot-aiding harvesting, each picker from a set $\mathcal{S}^{\mathcal{P}} = \{P_1, \ P_2, \ \ldots, P_Q\}$ of $Q$ pickers harvests fruits in a tray that lies on a picking cart. A team of $M$ identical transport robots $\mathcal{S}^{\mathcal{F}} = \{F_1, \ F_2, \ \ldots, \ F_M\}$ bring empty trays to the picker and carries the full tray to a collection station; the station's coordinates $\boldsymbol{L^s}$ are known. The robot scheduling algorithm has access to a set of predicted tray-transport requests $\mathcal{S}^{\mathcal{R}} = \{\mathcal{R}_1, \ \mathcal{R}_2, \ \ldots, \ \mathcal{R}_N\}$ where $0 \leq N \leq Q$.

Let us assume that at an instant $t_0$, $\mathcal{R}_i$ is different from the deterministic request that contains the following (known) information: (1) a prediction distribution of the remaining time interval $\aleph(\Delta t_i^f)$ with respect to $t_0$ until the tray becomes full of harvested fruit, (2) the predicted moving speed along the row $\aleph(v_i^y)$ while picking, and (3) the current location of the picker $\boldsymbol{L_i}$. $\aleph(\Delta t_i^f)$ is calculated from recent measurements from the load cells and $\aleph(v_i^y)$ is computed from recent GPS readings. The main methodology for building these predictions is explained in this work (Khosro Anjom & Vougioukas, 2019). The distribution of $\aleph(\Delta t_i^f)$ and $\aleph(v_i^y)$ followed Gaussian distributions. $\aleph(\Delta t_i^f)$ was achieved by linear regression model to predict the value of full tray time at the weight of the tray capacity. Mean of $\aleph(v_i^y)$ was obtained by linear regression to estimate the slope parameter and standard deviation of $\aleph(v_i^y)$ was obtained from the standard error of the regression coefficient.

A fast and near-optimal approach, MSA (Pillac et al., 2013), was adopted and adapted to incorporate the dynamic stochastic predictive requests in the computation of the schedule, assuming a limited computational power is available in this agricultural simulation. To implement MSA, two application-dependent functions must be setup: (1) a function GET-

SAMPLES $(\mathcal{S}^{\mathcal{R}}, M)$ which returns a set of M deterministic scenarios $\mathcal{S}^\xi = \{\mathcal{S}^{\xi_1}, \mathcal{S}^{\xi_2}, \ldots, \mathcal{S}^{\xi_M}\}$. Each scenario $\mathcal{S}^{\xi_i}$ contains a set of $N$ sampled requests. Each of the deterministic requests $R_i$ is sampled from predictive transport request distributions $\aleph(\Delta t_i^f), \aleph(v_i^y)$ of $\mathcal{R}_i$ in $\mathcal{S}^{\mathcal{R}}$; (2) a function OPTIMAL-SCHEDULE $(\mathcal{S}^{\xi_i})$ which returns an optimal schedule given a deterministic sampled scenario $\mathcal{S}^{\xi_i}$. The schedule includes the request rejections to some pickers and serving order for the remaining requests (3) a *consensus* function that combines all the individual scenario solutions into a single execution plan. For the function GET-SAMPLES $(\mathcal{S}^{\mathcal{R}}, M)$, the Monte Carlo sampling method was used to get the $M$ sampled scenarios from two distributions $\aleph(v_i^y)$ and $\aleph(\Delta t_i^f)$.

In a sampled scenario $\mathcal{S}^{\xi_i}$, each deterministic predictive request $R_i$ is composed of two sampled components, $\Delta t_i^f$ and $v_i^y$. Given them, a deterministic full tray location $\boldsymbol{L}_i^f$. can be calculated. Then, the variables relevant to the modeled scheduling problem can be calculated shown as Table 1 (Peng & Vougioukas, 2020).

*Table 1. Definitions of Symbols used in the modeling of deterministic predictive scheduling*

| | |
|---|---|
| $\boldsymbol{L}^s$: | the collection station location; |
| $\boldsymbol{L}_i^f$ | the full tray location in the field frame. |
| $D_{si}$: | one-way traveling distance, the Manhattan distance from $\boldsymbol{L}^s$ to $\boldsymbol{L}_i^f$ along the path; |
| $\Delta t_i^u$: | the corresponding robot's one-way travel time calculated by $D_{si}$ and robot speed; |
| $\Delta t^L$: | time interval when the picker takes the empty tray from the robot and loads the full tray on the robot (and then resumes picking); |
| $\Delta t^{UL}$: | the time interval for the collection station to unload the carried tray from the picker/robot and return an empty tray to the picker/robot; |
| $\Delta t_i^p$: | The total processing time required by a robot to serve request Ri and be available to serve another request; |
| $\Delta t_i^r$: | release delay of request Ri, the greatest value that eliminates robot idle time at $\boldsymbol{L}_i$. $\Delta t_i^r = max\big(\Delta t_i^f - \Delta t_i^u\big), 0\big)$ |

$\Delta t_k^A$:     The robot is available to be dispatched again, after a time interval $\Delta t_k^A$. $\Delta t_k^A = 0$, if the robot is available at the collection station;

$t_{ki}^d$:     The dispatch time instant of robot $F_k$ to the request $R_i$, which is no earlier than $t_0 + \Delta t_i^r$.

The pickers' requests may be rejected by the scheduler. In this case, they need to transport the full tray themselves and their self-transporting behavior is modeled following our previous work (Seyyedhasani et al., 2020b, 2020a). If the picker transports the tray themselves the total time $\Delta t_i^T$, required to deliver the full tray and take an empty tray back to resume picking is shown as (Eq 1). $\Delta t_i^{u_p}$ is the one-way travel time interval from full tray location $\boldsymbol{L}_i$ to $\boldsymbol{L}^s$ by the picker in $R_i$. $\Delta t_i^{u_p}$ is calculated based on $D_{si}$ and an estimated picker self-transport speed $v_p^i$ from the historic data of pickers. $\Delta t^{UL}$ is assumed to be constant depending on the crew management in the harvesting field.

$$\Delta t_i^T = 2\Delta t_i^{u_p} + \Delta t^{UL} \qquad \text{(Eq 1)}$$

The tray completion time instant, $t_i^{C_P}$ if the full tray is transported by the picker himself, is shown in (Eq 2) with $\Delta t_i^T$ representing the estimated tray-transport time by the picker.

$$t_i^{C_P} = t_i^f + \Delta t_i^T \qquad \text{(Eq 2)}$$

If the request is served by a robot $F_k$, the time instant, $t_i^{C_R}$ to resume picking is expressed as (Eq 3). The picker can start picking the next tray after the robot arrived at the full tray location and the full tray is exchanged with the empty tray from the robot.

$$t_i^{C_R} = t_{ik}^d + \Delta t_i^u + \Delta t^L \qquad \text{(Eq 3)}$$

The nonproductive time, $\Delta t_i^N$ of $R_i$ can be calculated as (Eq 4). The objective of the modeled problem is to minimize the mean of the nonproductive time of all the pickers. In the objective function, both $t_i^{C_P}$ and $t_i^{C_R}$ are represented by $t_i^C$ which is decided by the decision variables.

$$\Delta t_i^N = t_i^C - \Delta t_i^f \qquad \text{(Eq 4)}$$

After building the mathematic equations of these modeled variables, the function OPTIMAL-SCHEDULE ($\mathcal{S}^{\xi_i}$) was built to find the solution for each scenario. First, the exact solution is computed using integer programming to get the best possible solution. Second, a fast and sub-optimal heuristic policy is implemented to get a near-optimal solution in less time, so that the pickers do not need wait for long time caused by the scheduling computation. The results of the exact and heuristic solutions are compared in the proposed performance metrics in Section 5.

### 4.1. Scenario solution using integer programming

The deterministic predictive scheduling problem of each sampled scenario was modelled using an integer linear program. $\mathcal{S}^{\mathcal{T}}$ is used to represent the discretized time set, $\{1, 2, 3..., TB\}$. $TB$ is the upper bound makespan of all requests (from $t_0$ to $t_0 + max_i\{t_i^C\}$). For this problem, the upper bound $TB$ can be expressed as (Eq 5). It is easy to prove that the completion time of any request cannot be larger than the maximum completion time of self-transporting, otherwise that request should be transported by the picker themselves.

$$TB \le t_0 + max_i\{\Delta t_i^f\} + max_i\{\Delta t_i^T\} \qquad \text{(Eq 5)}$$

The decision variable is defined as $\chi_{ikt}$, where $i$ is the index of request $R_i$. $k$ is the index of the serving robot if $1 \le k \le M$; $k = M + 1$ means that the picker transports the tray himself. $t$ is the index of discrete-time instant. $\chi_{ikt}$ is equal to 1 if $R_i$ is served by a robot $F_k$ ($1 \le k \le M$) or transported by the picker himself ($k = M + 1$) at the time instant $t$. The problem can be modeled using an integer linear programming (ILP) as follows.

$$min \sum_{i=1}^{N} \Delta t_i^N$$

$s.t.$

$$\sum_{k=1}^{M} \sum_{t=1}^{\Delta t_i^T} \chi_{ikt} = 0, \ R_i \in \mathcal{S}^{\mathcal{R}} \tag{Eq 6}$$

$$\sum_{i=1}^{N} \sum_{t=1}^{\Delta t_k^A} \chi_{ikt} = 0, \ 1 \le k \le M \tag{Eq 7}$$

$$\sum_{k=1}^{M} \sum_{t=1}^{t_i^f} \chi_{i(M+1)t} = 0, \ R_i \in \mathcal{S}^{\mathcal{R}} \tag{Eq 8}$$

$$\sum_{k=1}^{M+1} \sum_{t=1}^{TB} \chi_{ikt} = 1, \ R_i \in \mathcal{S}^{\mathcal{R}} \tag{Eq 9}$$

$$\sum_{k=1}^{M} \sum_{t=max(1,t-\Delta t_i^p)}^{t} \chi_{ikt} \le 0, \ R_i \in \mathcal{S}^{\mathcal{R}} \tag{Eq 10}$$

$$t_i^C = \sum_{k=1}^{M} \sum_{t=1}^{TB} (t + t_i^U + \Delta t^L)\chi_{ikt} \ , \ R_i \in \mathcal{S}^{\mathcal{R}} \tag{Eq 11}$$

$$t_i^C = \sum_{t=1}^{TB} (t + t_i^T)\chi_{i(M+1)t} \ , \ R_i \in \mathcal{S}^{\mathcal{R}} \tag{Eq 12}$$

$$\Delta t_i^N = t_i^C - \Delta t_i^f, \ R_i \in \mathcal{S}^{\mathcal{R}} \tag{Eq 13}$$

The objective function is the sum of the non-productive time of all requests and the required constraints are explained as follows. In (Eq 6), it represents that any request cannot be served by a robot before their release constraints. (Eq 7) means that the robot's start serving time cannot be earlier than their initial available time. If the request is transported by the picker himself, the start time cannot be earlier than the full tray instant $t_i^f$ as (Eq 8). (Eq 9) represents that all requests must be served either by a robot or by the picker himself. (Eq 10) shows that any requests can be served by only one robot (preemption is not allowed). (Eq 11) expresses the tray completion time of the request served by the robots, while (Eq 12) is the tray completion time served by the pickers themselves. (Eq 13) shows the non-productive time of request $R_i$.

As mentioned above, predictive scheduling of crop-transport robots is a variant of the Parallel Machine Scheduling Problem (PMSP). Following symbol notations defined by Lawler et al. (1993), the problem is referred to as $Pm|r_i|\sum C_i$, where $Pm$ represents identical parallel machines, means that the $i$th $r_i$⌷ represents $\sum C_i$ shown that this problem is NP-hard in a strong sense and hence the optimal solution cannot be obtained in polynomial time ⌷⌷ In this paper, the modeled ILP was solved by a commercial solver (Gurobi Optimization, LLC., 2020) at the cost of long computation.⌷

## 4.2.    Scenario solution with heuristic policy

A heuristic policy, namely, the shortest release time with long process time first (SRLPT), is proposed to achieve a fast but sub-optimal result in each sampled deterministic scenario. The requests reaching the release constraint ($\Delta t_i^r = 0$) will enter a scheduling pool and the request with the longest process time in the pool is ordered to be served by the first available robot. The non-productive time of those requests with large self-transport time can be reduced significantly by the service of available robots. The requests with a shorter transport time are served late, as even if they are rejected, the non-productive time will not be that large. The requests with a full tray location less than 5 meters away from the end of the row are rejected, as those pickers only need to walk a small distance back and forth to resume picking. The performance comparison between the heuristic policy and ILP is shown in Section 5.2.

## 4.3.    Consensus function

Given solutions in multiple sampled scenarios, the consensus function is to select the distinguished plan from the current pool of scheduling plans. Bent and Pascal (2004) first applied the consensus function into a modeled partially dynamic vehicle routing problem with a time window. They pointed out that this approach is essentially domain independent. Hence, they applied a similar consensus approach on the classic scheduling problem, the online packet scheduling problem in computer networks. The key idea is to solve each sampled scenario once and to select the packet which is most often in the optimal solution of each scenario. The heuristic idea behind the consensus function is the least-commitment approach, a well-known approach in the artificial intelligence community (R. Bent & Van Hentenryck, 2004 June). By choosing the job that occurs the most often, the consensus algorithm takes a decision that is consistent with the optimal solution of many samples.

This consensus approach was applied to our modeled scheduling problem. After all the deterministic scenarios are solved with the function of OPTIMAL-SCHEDULE, the scheduling plan of each scenario is converted to a serving order based on their scheduled serving times. A score function is defined for each request in one scenario. If the request is rejected, the score value of that request is counted as -1. If the request is served by a robot in the order of $O_i$ among all the serving requests in that scenario, the score of that request is counted as $(N - O_i)$. The score of each request is obtained by adding the scores of the requests among all the sampled scenarios. The consensus serving order is the descending order of the scores of all the requests in $\mathcal{S}^{\mathcal{R}}$. The available robots were dispatched to the first request in the consensus order at the instant when the expected release time of that request is reached. The rejection flags are sent to the pickers if they are not served by the robot at the instant when their trays are full. The scheduler will run to update the scheduling plan only when there are robots available and new transport requests entering the set.

# 5. Experiments and results

## 5.1. Evaluation metrics

In both the all-manual and robot-aided strawberry harvesting trials, the productive time per tray – denoted as $\Delta t_i^{ef}$ – is defined in the same way: it is the time required by a picker to fill the $i$th tray to its capacity, starting from an empty tray. Productive time includes picking and walking to relocate to a new furrow to resume picking when the tray cannot be finished in the current furrow. Non-productive time per tray – denoted as $\Delta t_i^{fe}$ – is defined as the time interval that is not spent picking or relocating to pick from another furrow. In manual strawberry harvesting, $\Delta t_i^{fe}$ includes the picker's walking time to transport the full tray to the unloading station, the waiting time in a queue to deliver the tray and get an empty one, and the walking time required to return to the previous position to resume picking. In contrast, in robot-aided harvesting, $\Delta t_i^{fe}$ is the sum of the time, $\Delta t_i^w$, the picker spends waiting for a robot to arrive, plus the time, $\Delta t^L$, needed to place the full tray on the robot and take an empty tray from the robot. $\Delta t_i^w$ is highly dependent on the robot scheduling policy, whereas $\Delta t^L$ is small and is assumed to be constant. $\Delta T^{fe}$ represents the average non-productive time of all the trays measured in an experiment.

The mean harvesting efficiency, $E_{ff}$, when harvesting N trays with or without robots, is defined as the averaged sum of ratios of productive time over total time spent for each tray; it is calculated by (Eq 14):

$$E_{ff} = \frac{1}{N} \sum_{i=1}^{N} \frac{\Delta t_i^{ef}}{\Delta t_i^{ef} + \Delta t_i^{fe}} \quad \text{(Eq 14)}$$

$\Delta T^{fe}$ and $E_{ff}$ can be used to evaluate the overall performance of all-manual and robot-aided harvesting.

## 5.2. Simulation experiments

The simulation experiments had two goals. The first goal was to compare the performance difference between the heuristic (SRLPT) and exact (ILP) optimization algorithms introduced in Section 4. The second goal saw to select the number of scenarios that the MSA will sample when executing in real-time, during field experiments. It is known that when more scenarios are sampled, the MSA solution improves, but takes a longer time to compute; hence, a tradeoff must be reached.

The simulation experiments were performed using a Monto-Carlo simulator that was adapted from our previous work (Peng & Vougioukas, 2020). The robot speed was 0.4m/s on the headland and 1.2 m/s inside the furrows, and the FR threshold was 0.7. Each harvesting scenario was simulated by running 100 Monte-Carlo runs, as described in that work. The main underlying assumption in our analyses of the results, is that the 100 sampled means of each evaluated metric were normally distributed. The random parameters of the requests were generated following the distributions measured by Khosro Anjom et al. (2019): the mean of bias for the full tray time prediction is less than 10% of one tray picking time, and the standard error of the prediction is 30 s.

Toward the first goal, the ILP and SRLPT single-scenario optimal scheduling solvers were implemented in MSA's OPTIMAL-SCHEDULE module and compared for an increasing number of robots (4 to 12) and a typical crew size of 25 pickers. Toward the second goal, the harvesting efficiency achieved with the MSA was computed for an increasing number of scenarios (1 to 80), with eight robots and 25 pickers. Different pairs of numbers of scenarios (e.g., 30 vs. 50) results were compared using Tukey's Honest Significant Difference (HSD) tests, to choose the number of scenarios beyond which performance did not improve significantly.

### 5.2.1. Adaption of FSM in the hybrid simulator

In our previous work (Peng & Vougioukas, 2020), a discrete-time hybrid systems model was developed to model and simulate the activities and motions of all agents involved in robot-aided harvesting. A Finite State Machine (FSM) was utilized to model the discrete operating states/modes of the agents and the transitions between the operating modes. In this work, the FSM was extended to incorporate the changes in the scheduling algorithm (stochastic transport requests and tray-transport request rejections). The activities of a picker during robot-aided harvesting were classified into 14 discrete operating states/modes (Table 2), and the operations of a tray-transport robot into 9 states (Table 3). The operating states of pickers and robots and the possible transitions amongst them are shown in Figure 17. Next, the changes in the picker and robot FSMs are described in detail.

In the picker FSM, pickers need to transport the trays themselves if they receive the request rejections. In FSM of robots, it may happen that the robot is dispatched to a row where the served picker cannot fill their tray and take the half-filled tray to the next unharvested row. In this case, the robot drives back to the collection station to wait for the next dispatching command.

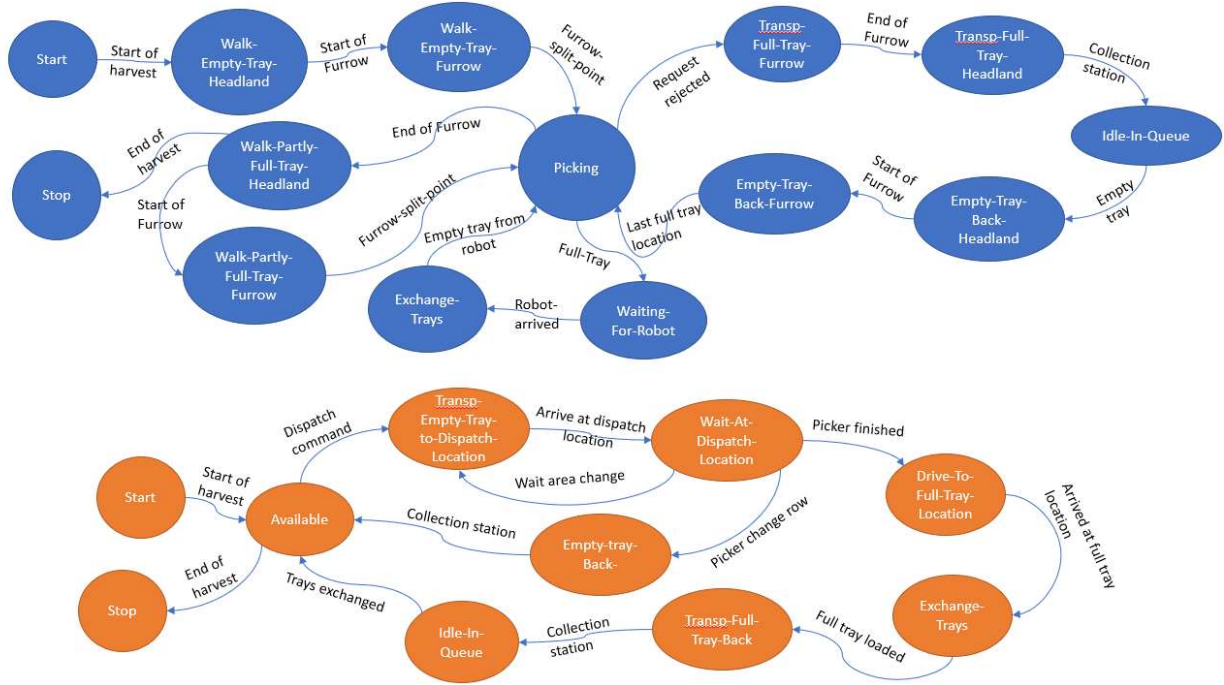

*Figure 17. State diagram of picker states and transport robot states during human-robot collaborative harvesting*

*Table 2. States defined to represent a picker's operating states during robot-aided harvesting*

| Operating state | Action |
| --- | --- |
| Start | A picker leaves the collection station with an empty tray in hand, to start picking. |
| Walk-Empty-Tray-Headland | A picker walks with an empty tray on the headland, toward an empty (unoccupied) furrow. |
| WALK-Empty-Tray-Furrow | A picker walks inside an empty (unoccupied) furrow with an empty tray until the field's split line is reached. |
| Picking | A picker is picking inside a furrow, with direction from the field split line toward the collection station. |
| Waiting-For-Robot | A picker waits (idle), with a full tray, for a robot to come. |
| Exchange-Trays | A picker takes the empty tray brought by the robot and places a full tray on the robot. |
| Walk-Partly-Full-Tray-Headland | A picker takes partly full tray on the headland, toward an empty (unoccupied) furrow. |
| Walk-Partly-Full-Tray-Furrow | A picker takes a partly full tray inside an empty (unoccupied) furrow until the field's split line is reached. |
| Transport-Full-Tray-Furrow | A picker takes a full tray inside a furrow towards the headland |

| Transport -Full-Tray-Headland | A picker takes a full tray on the headland towards the collection station |
|---|---|
| Idle-In-Queue | A picker waits in a line at the collection station to deliver her/his full tray and receive an empty tray. |
| Empty-Tray-Back-Headland | A picker walks in the headland - toward the last full tray furrow - carrying an empty tray, to continue harvesting. |
| Empty-Tray-Back-Furrow | A picker walks back to the last full tray location with an empty tray, to continue harvesting. |
| STOP | A picker stops picking after the last tray is picked up by a robot. |

*Table 3. States defined to represent a robot's operating states during robot-aided harvesting*

| Operating state | Action |
|---|---|
| Start | A robot at the collection station starts operation with no tray on it. |
| Available | A robot with one empty tray on it is waiting at the collection station to be dispatched to a tray-transport request. |
| Transp-Empty-Tray-to-Dispatch-Location | A Robot travels from a collection station – carrying an empty tray – toward the dispatched location. |
| Wait-At-Dispatch-Location | A robot arrives at the location of the tray-transport request and waits for the picker to finish harvesting. |
| Drive-To-Full-Tray-Location | A robot drives to picker's full tray location after served picker fills the full tray in its dispatched row. |
| Empty-Tray-Back | A robot runs back to collection stations as the served picker cannot fill the tray in its dispatched row |
| Exchange-Trays | A robot is idle while the picker exchanges the empty tray with a full tray. |
| Transp-Full-Tray-Back | A robot travels toward the collection station to deliver a full tray. |
| Idle-In-Queue | A robot with a full tray waits in a queue at the collection station to have its tray unloaded, and an empty tray loaded. |
| Stop | A robot stops its operation at the collection station after the last tray has been unloaded. |

Request rejections are integrated into the operation of pickers. When the rejection flags are received, the pickers will transport the full tray by themselves as in manual harvesting. The relevant pickers' states are "Transport-Full-Tray-Furrow", "Transport -Full-Tray-Headland", "Idle-In-Queue", "Empty-Tray-Back-Headland" and "Empty-Tray-Back-Furrow". If the picker is served by a scheduled robot, they will wait at their full tray locations to exchange trays from the coming robot. In this case, the states after picking are "Waiting-For-Robot" and "Exchange-

Tray". The time spent between the full tray instant, and the starting instant of next tray picking is denoted as "non-productive" time.

The rest of the picker states and the dynamics inside each state are the same as in our previous work (Seyyedhasani, Peng, Jang & Vougioukas, 2020a). The stochastic parameters were estimated experimentally, as in (Peng & Vougioukas, 2020).

Since the predictions of tray-transport requests contain uncertainty, the exact location when the tray will become full is not known exactly. Therefore, instead of sending the robot to the predicted location, a safety distance of 5 meters from the predicted full-tray location was introduced for the robots' goal points. The pickers need to walk this small distance to load the tray onto the robot and take an empty tray back. The other states of the crop-transport robots are updated as in (Peng & Vougioukas, 2020).

### 5.2.2. Simulation results and analysis

We first evaluated the performance of the proposed heuristic policy (SRLPT) and ILP policy given deterministic requests (Peng & Vougioukas, 2020). The comparison of evaluation metrics $E_{ff}$ is shown in Figure 18 for an increasing number of robots.

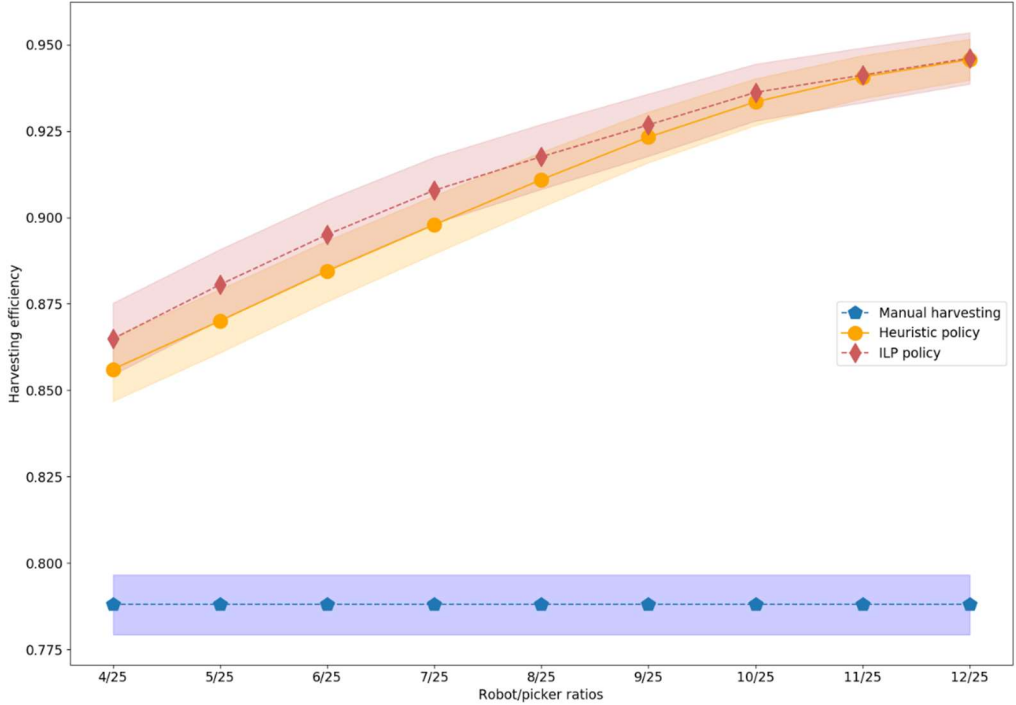

*Figure 18. Mean values of harvesting efficiency (points) and their 95% CI (shaded areas) as a function of the number of robots for the scheduler with ILP and heuristic SRLPT.*

T-tests were applied for the mean efficiencies of the two policies under different robot-picker ratios. The null-hypothesis of each T-test was that the mean efficiencies of the two policies were the same. The alpha value for the tests was set to 5% (Type I error). The results are presented in Table 4, where one can see that the performance of two policies was not significantly different when the robot/picker ratio was over 8/25, as all p values of the T-test results were over 0.05.

*Table 4. P values of T-tests for the efficiencies of two policies in different robot/picker ratios*

| Robot/picker ratio | 4/25 | 5/25 | 6/25 | 7/25 | 8/25 | 9/25 | 10/25 | 11/25 | 12/25 |
|---|---|---|---|---|---|---|---|---|---|
| P values | 0.030 | 0.015 | 0.013 | 0.024 | 0.055 | 0.118 | 0.140 | 0.379 | 0.488 |

Then, we investigated the performance of the MSA scheduler as a function of the number of sampled scenarios, given the experimentally measured distribution of the request prediction uncertainty from the work of Khosro Anjom et al. (2019). The mean values of harvesting efficiency and their 95% CI (shaded areas) as a function of the number of sampling scenarios are shown in Figure 19.

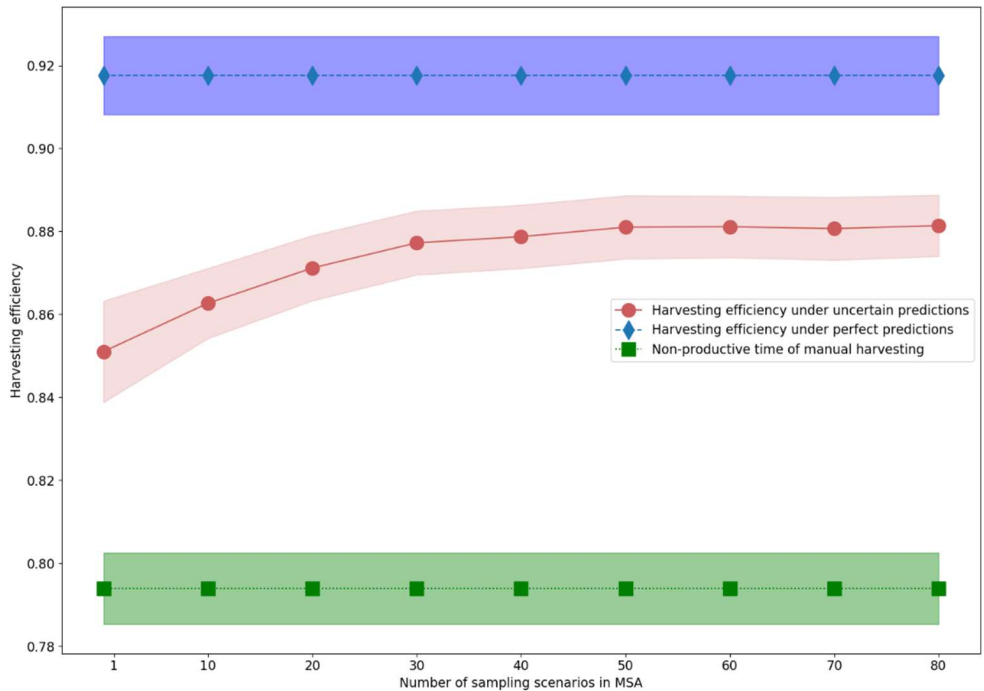

From the results, one can see that as the number of scenarios in MSA increases, the mean harvesting efficiency (red curve) increases when the number of scenarios is less than 50. However, more than 50 scenarios are considered, the efficiency and non-productive time come to a plateau. Tukey's HSD tests were used for different sampling scenarios {10,20,30,40,50} to examine their efficiency differences, which is presented in Table 5. The null hypothesis was that the mean efficiency of MSA in different sampling scenarios were the same. The alpha value for the tests was set to 5% (Type I error). From the combinations of {30, 40}, {30, 50} and {40, 50} in Table 5, one can see that the scheduling performance of MSA did not show significant improvement when the number sampling scenarios was over 30.

*Table 5. Tukey's HSD results of harvesting efficiencies for different sampling scenarios in MSA*

| Scenarios comparing combinations | | Adjusted p-values | Null rejections |
|---|---|---|---|
| 10 | 20 | 0.0012 | True |
| 10 | 30 | 0.0013 | True |
| 10 | 40 | 0.0010 | True |
| 10 | 50 | 0.0005 | True |
| 20 | 30 | 0.0022 | True |
| 20 | 40 | 0.0031 | True |
| 20 | 50 | 0.0020 | True |
| 30 | 40 | 0.7631 | False |
| 30 | 50 | 0.0752 | False |
| 40 | 50 | 0.1221 | False |

However, the mean of computation time for a schedule with 50 scenarios is approximately 5 seconds (on the Intel Core i7-3770@3.40 GHZ laptop used as a server), which is adequate for real-time operation. As a result, we set the number of sampling scenarios for the MSA to be 50.

## 5.3. Field experiment design

The main goal of the field experiments was to evaluate the savings of the harvest-aiding system in commercial strawberry harvesting, with a crew of professional pickers. On Nov 10[th] and Nov 11[th], 2020, the harvest-aiding system was evaluated with a crew of six professional pickers in a commercial field near Lompoc, CA shown as Figure 20. All the field experiments were conducted under the UC Davis Institutional Review Board (IRB) compliance protocol "IRB 575389-8". Each day, the pickers' working schedule was divided into 2 sessions. The first session was from 8:00 am to 11:00 am and the second session was from 11:30 am to 2:30 pm. In the first session, on November 10th, our system was set up and tested on the mapped field while the crew harvested the orange-colored field block (Figure 20b) in their usual manner, using our instrumented carts. The pickers collected the strawberries in 500-gram carton box trays. The gross mass of a full tray was around 4.5 Kg (10 lbs). Their harvesting data was collected and saved on the SD card modules of the carts. From the data in that session, their walking speeds when transporting full trays were estimated. This data was used to estimate the performance of manual harvesting.

In the second session of November 10th and the first session of Nov 11th, all 6 pickers started harvesting from the field's middle line and moved toward the unloading station in their typical harvesting manner. The crew harvested with the assistance of two robots (blue area for Nov 10th and red area for Nov 11th on Figure 20b). The harvesting data using the co-robotic harvest-aiding system were recorded into ROS files on the server laptop, as well as in the SD cards of the carts. The evaluation results of our harvest-aiding co-robotic system (Section 2.2) were obtained from these two sessions.

During robot-aided harvesting the pickers were asked to press the request button on the cart once they filled 6 out of the 8 clamshell boxes in their tray. It was explained to them that if their transport request was accepted by the robots, the yellow LED on their cart would turn on. In this case, they were instructed to wait for a robot, in case they filled their tray and a robot had not arrived. The scheduling system dispatched robots to serve the requests by solving online the stochastic predictive scheduling problem. The robots were scheduled and dispatched to the predicted full-tray locations inside the rows. Upon arrival at the commanded location, the robots would stay still until the picker placed their tray on the robot and pressed a button that sent the robot back to the collection station.

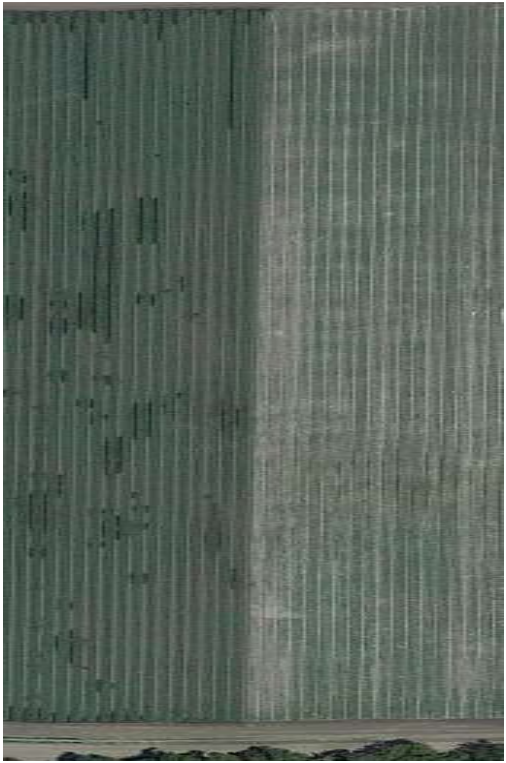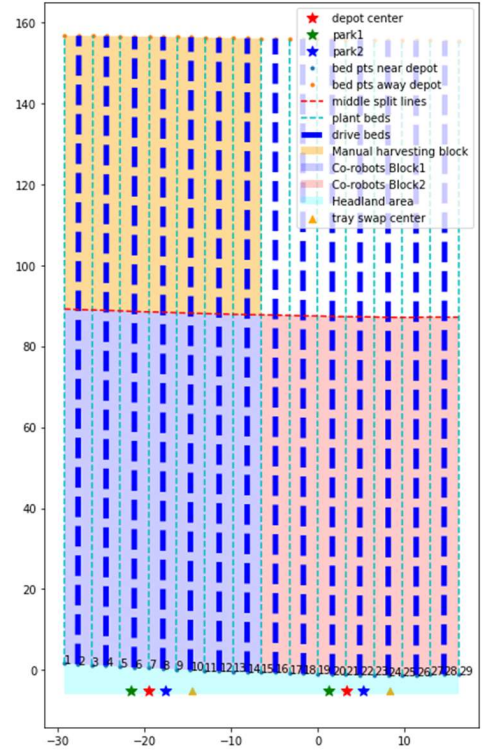

*Figure 20. a) Satellite picture of commercial field block near Lompoc, CA from Google Maps; b). Map of the field block built with RTK: blue and orange shaded areas were evaluated on Nov 10th; red and white shaded areas were evaluated on Nov 11th*

The system components are shown in Figure 21. The full-empty tray swap location was approximately 5 meters away from the depot center where the server laptop was located (red star in Figure 20.a), on its right-hand side, facing the field.

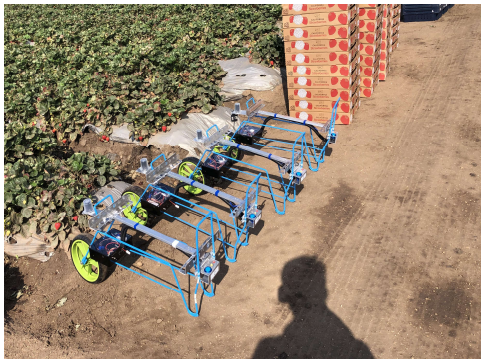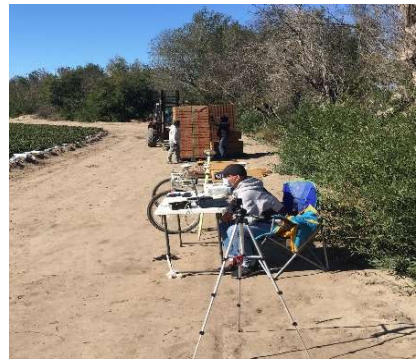

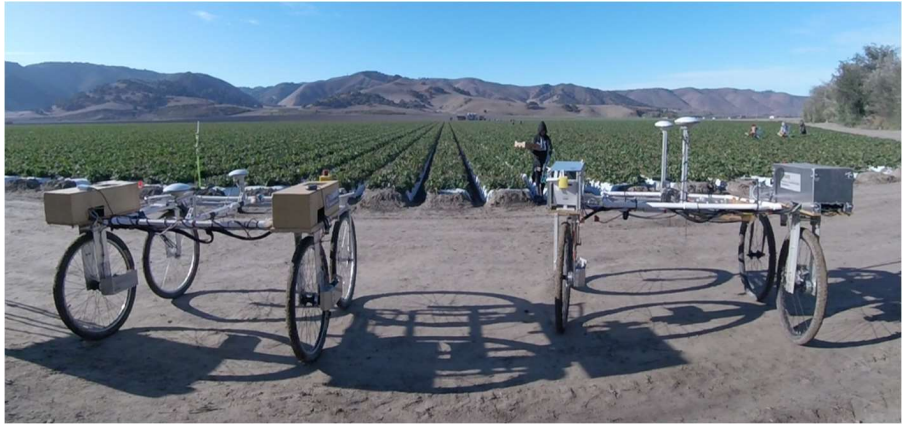

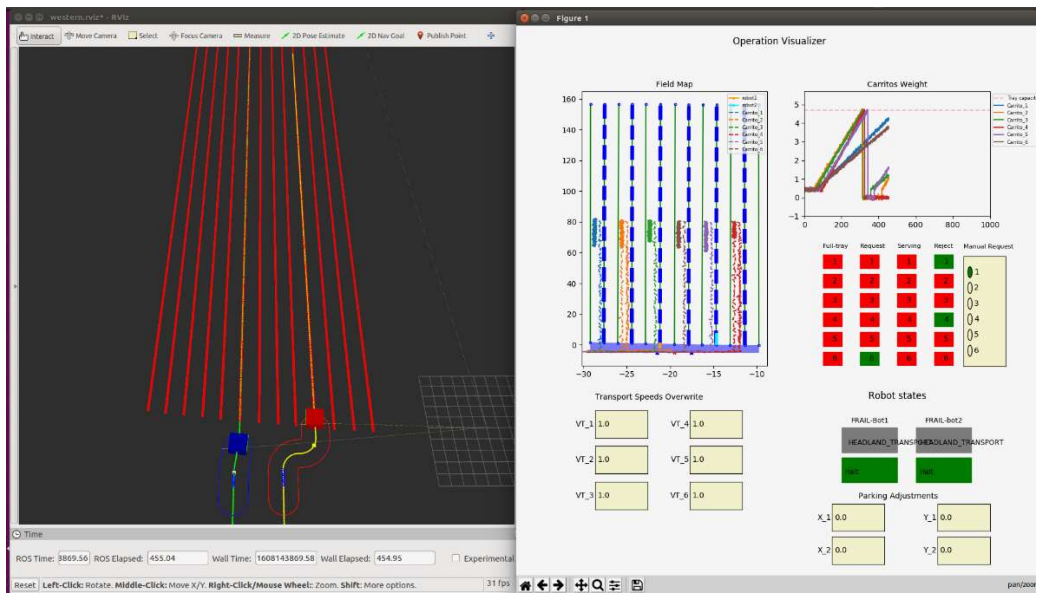

*Figure 21. Components of robotic harvest-aiding system: a) Instrumented carts; b) Collection station (aka, depot center) with scheduling server; c) Two FRAIL-Bots parked at the depot center; d) Scheduling server user interface built with Python Matplotlib and ROS RVIS for visualization and monitoring of the robots motions*

In the first session on Nov 10[th], 33 trays of fruit were harvested manually in the field block. In the second session on Nov 10[th], 41 trays were harvested by the 6 pickers working with the co-robotic harvest-aiding system. In the first session on Nov 11[th], the same picking crew of 6 people harvested 24 trays of fruit using the harvest-aiding system. The data from these three sessions was processed and analyzed in Section 5.4. Non-parametric tests were used to compare the performance of manual and co-robotic harvesting.

## 5.4. Experimental data processing

The processing of the collected data is described in this section. Table 6 shows the data fields recorded on the SD cards of the instrumented carts and transmitted wirelessly to the LoRa server. On the SD card, the data was recorded at the GPS epoch update frequency, i.e., at 10 HZ. The LoRa message transmission frequency of each cart was 1.6 Hz.

*Table 6 Data protocol recorded in the SD card on the instrumented cart*

| Item | Time stamp | Longitude | Latitude | Filtered mass | Button state |
|---|---|---|---|---|---|
| Description | GPS epoch with accuracy of 100ms | Geodetic coordinates of the cart location | | Filtered mass readings of the tray by the instrumented cart | Status of button on the cart: 0 represents pressed and 1 represents not pressed |
| Data Type (Arduino Due) | Unsigned Int (32 bits) | Double (64 bits) | Double (64 bits) | Float (32 bits) | Unsigned short (8 bits) |

After receiving and processing a LoRa message, the "Cart-states-pub node" (Figure 14) converted it into a ROS message whose contents are shown in Table 7. The geodetic coordinates of the cart were converted to the coordinates in the field map.

*Table 7. Cart messages in ROS, converted by the LoRa server ROS node*

| Item | Header | Cart's ID | X | Y | Button Request |
|---|---|---|---|---|---|
| Description | Time stamp of the data received. Frame ID of the data (local field frame); | 1~6 | Coordinate of the cart in the field map | | Request of the picker: 1 represents requesting and 0 represents not requesting |
| Data type (ROS) | std_msgs/Header.msg | std_msgs/Byte | std_msgs/Float32 | | std_msgs/Byte |

The methodology described in previous work (Khosro Anjom & Vougioukas, 2019) was used to compute these metrics from the cart and picker data; the time instants shown in Figure 22 were extracted from the logged data. The time instant $t_i^{\{end\}}$ when picking of a tray ended was identified by detecting a big drop in the measured mass of the tray, after the measured mass exceeded 4,000 grams. The instant $t_i^{\{start\}}$ that picking started was found by detecting the moment when the measured mass was around 500 grams. A tray-transport request time instant corresponded to a change in the state of the request button from "0" to "1". After extracting $t_i^{\{start\}}$ and $t_i^{\{end\}}$ for each tray, the productive interval $\Delta t_i^{ef}$ was calculated by (Eq 15) and the non-productive interval $\Delta t_i^{fe}$ was calculated using (Eq 16). The efficiency of the tray was calculated by (Eq 17).

$$\Delta t_i^{ef} = t_i^{\{end\}} - t_i^{\{start\}} \qquad \text{(Eq 15)}$$

$$\Delta t_i^{fe} = t_{i+1}^{\{start\}} - t_i^{\{end\}} \qquad \text{(Eq 16)}$$

$$\mathrm{E_{ff}}_i = \frac{\Delta t_i^{ef}}{\Delta t_i^{ef} + \Delta t_i^{fe}} \qquad \text{(Eq 17)}$$

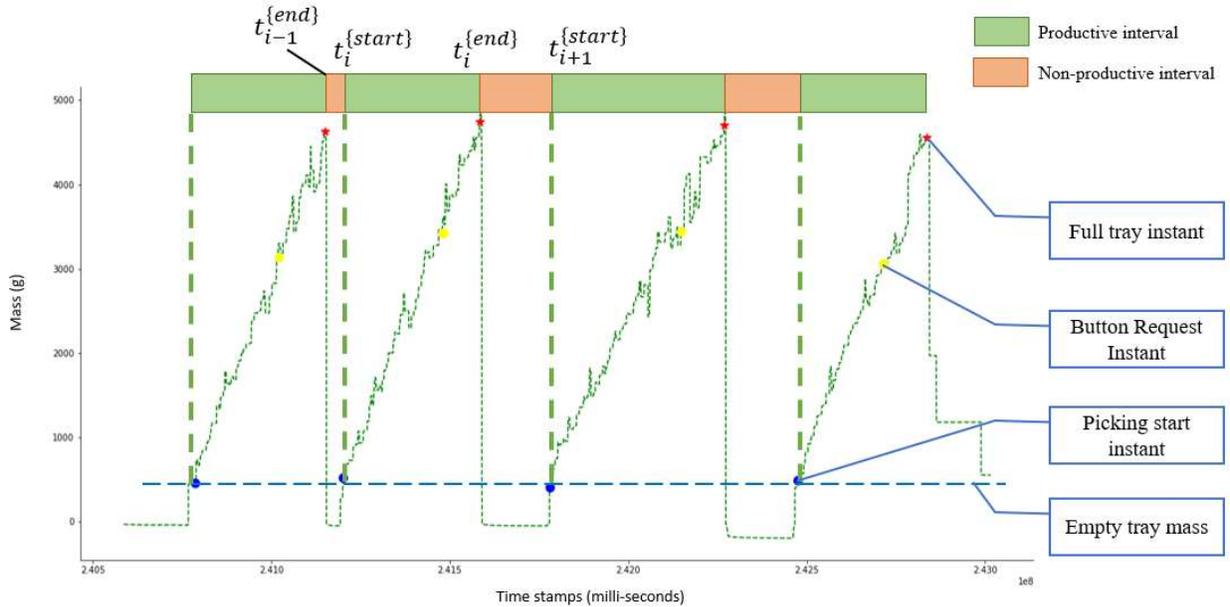

*Figure 22. Visualization of tray mass measurements and time stamps from the collected data: blue dots represent picking start points; yellow dots represent tray-request events (button pressed); red stars represent picking end points.*

## 5.5. Analysis of results

The data from each instrumented cart was processed with the method described in Section 5.3 to compute the productive time, $\Delta t_i^{ef}$, (aka 'one-tray-picking time') and non-productive time, $\Delta t_i^{fe}$, of each tray. The distance traversed to pick a single tray ('one-tray-picking distance') was obtained by calculating the Euclidean distance of the cart locations at the instances when picking started and ended for the tray. The results were combined and presented in this section.

### 5.5.1. Results of harvesting parameters

The distributions of the one-tray-picking time, one-tray picking distance, and picker walking speed parameters were generated from the data. The single-tray picking time and single-tray picking distance parameters were assumed to be dependent on the geospatial fruit distribution, for the same picking crew.

Figure 23 shows the distributions of the one-tray picking time of the crew for the co-robotic harvesting blocks of Nov 10[th] (session 2) and Nov 11[th] (session 1). The Mann-Whitney rank test was used to compare the two distributions. The null hypothesis was that for randomly selected values from the distribution of harvesting time per tray on Nov 10[th] and Nov 11[th], the probability of the selected values on Nov 10[th] being greater than Nov 11[th] was equal to the probability of selected values on Nov 11[th] being greater than Nov 10[th]. The significance level of p-values (alpha) for rejecting the null hypothesis (Type I error) was chosen as 1%. The calculated p value was 2.13e-9, so the distributions of the two days differed significantly. A comparison of their mean values shows that, on average, the harvest crew took a longer time to harvest one tray on Nov 11[th] (894.62 s) than on Nov 10[th] (548.71 s).

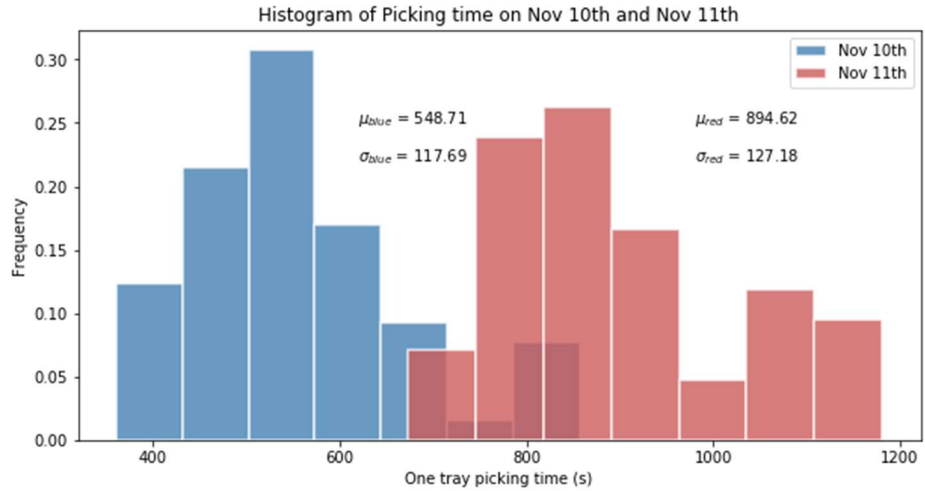

*Figure 23.  Histogram of the time it took pickers to fill one tray, on Nov 10$^{th}$ and Nov 11$^{th}$*

Figure 24 shows the distributions of the one-tray picking distance, for the two days. The p value from the Mann-Whitney rank test of the two distributions was 2.21e-7, so they were significantly different. On average, the pickers moved a longer distance to collect a tray of strawberries on Nov 11$^{th}$ (33.31 m) than Nov 10$^{th}$ (17.92 m).

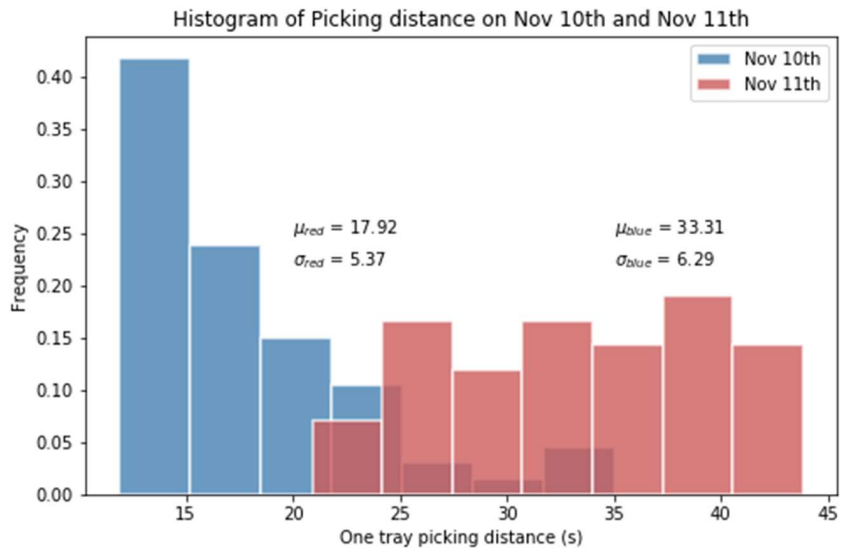

*Figure 24 Histogram of the distance traveled by pickers to fill one-tray, on Nov 10$^{th}$ and Nov 11$^{th}$*

During manual harvesting, each picker would take their filled tray to the collection station, attach a sticker with their personal barcode on the tray, take an empty tray and walk back to the field to resume picking. Based on our observations, the pickers took around 8 seconds to stick their bar code on the tray and take an empty tray. Thus, the walking time to deliver a tray can be estimated by subtracting these 8 seconds from $\Delta t_i^{fe}$. The exact locations $\boldsymbol{L}_i^f$ when a picker starts walking are detected by the weight change; the corresponding time instants are $t_i^{\{end\}}$. The transport distance was computed from the coordinates of the collection station and $\boldsymbol{L}_i^f$. Hence, each picker's walking speed was estimated from the computed transport distance and the measured time interval. The manual harvesting data from the first session of Nov 10th was used to estimate the mean walking speeds of the 6 pickers, as shown in Table 8.

*Table 8. The estimated mean walking speeds of the pickers, when they transport full trays to the collection station*

| Picker ID# | Sample mean of walking speed (m/s) | Sample standard deviation of walking speed (m/s) | Number of manual transport measurements |
|---|---|---|---|
| 1 | 0.78 | 0.05 | 5 |
| 2 | 0.49 | 0.10 | 5 |
| 3 | 0.81 | 0.06 | 6 |
| 4 | 0.74 | 0.11 | 5 |
| 5 | 0.91 | 0.07 | 6 |
| 6 | 1.02 | 0.03 | 6 |

### 5.5.2. Co-robotic harvesting performance

The non-productive time $\Delta t_i^{fe}$ and the efficiency $E_{ff_i}$ for each tray was obtained from the data collected during the co-robotic harvesting sessions. Obviously, it is impossible to have the picking crew re-harvest manually a field block that was harvested using the robots. Ideally, a large trial would use a large field and divide it into smaller blocks and then randomly assign manual and robotic treatments to the blocks. However, such a large trial was not acceptable by the grower; commercial harvesting is a costly operation that is planned based on the weather, crop condition, labor availability and customer demand. Hence, an estimation of the manual harvesting efficiency was made for the same field block that was harvested using robots, given the pickers' estimated walking speed and the measured locations where their trays had filled up.

As mentioned above, the location $L_i^f$ of each tray can be indexed from the instant $t_i^{\{end\}}$ the tray becomes full. Thus, the non-productive time and the manual harvesting efficiency for each tray were estimated from $L_i^f$, the collection station position and the estimated walking speeds (from Section 2.1), In summary, the picking crew's harvesting performance with the co-robotic harvest-aiding system was measured directly, and the manual harvesting performance of the same picking crew for the same field block was estimated indirectly, using the above method.

Given the significant difference in the yields of the two fields, which manifested itself in very different one-tray picking time and distance statistics, the harvesting performance was evaluated on the two days separately. The frequency histograms of non-productive time for the manual harvesting and co-robotic harvesting on Nov 10th are shown in Figure 25.a, and the histograms of harvesting efficiency on that day are shown in Figure 25.b. The P-value of Mann-Whitney testing results of the performance data (non-productive time and harvesting efficiency) of manual harvesting and co-robotic harvesting on that day are shown in Table 9. Based on the calculated P-values, the manual and co-robotic distributions of the non-productive time and efficiency are significantly different, with a significance level at 1%.

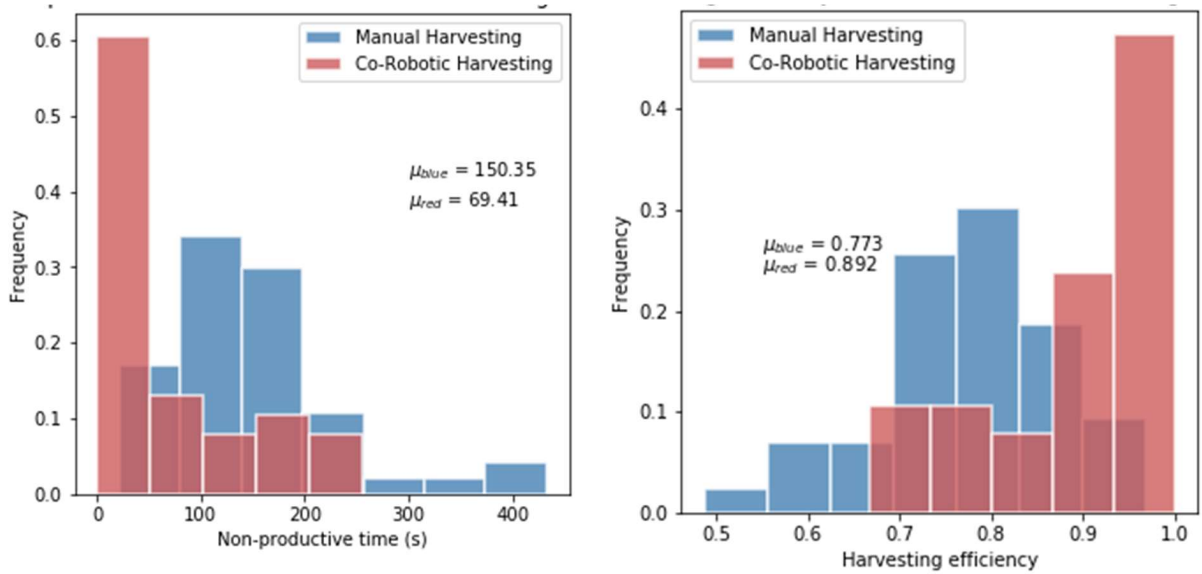

*Figure 25. Harvesting performance on Nov 10th: a) Histogram of non-productive time of Co-robotic and manual harvesting; b) Histogram of harvesting efficiency of Co-robotic and manual harvesting.*

*Table 9 Mann-Whitney rank test results for the means of the measured and estimated non-productive time and harvesting efficiency of the co-robotic and manual harvesting, respectively, on Nov 10th.*

| Item | Mean Non-productive time | Mean Harvesting efficiency |
|------|--------------------------|----------------------------|
| P value | 3.778e-6 | 1.3705e-6 |

Their mean values are shown in the figure. The mean non-productive time of co-robotic harvesting was reduced by more than half of the manual harvest non-productive time. The mean co-robotic harvesting efficiency increased by around 12% compared to manual harvesting. These results are shown in the corresponding pie charts, in Figure 26.

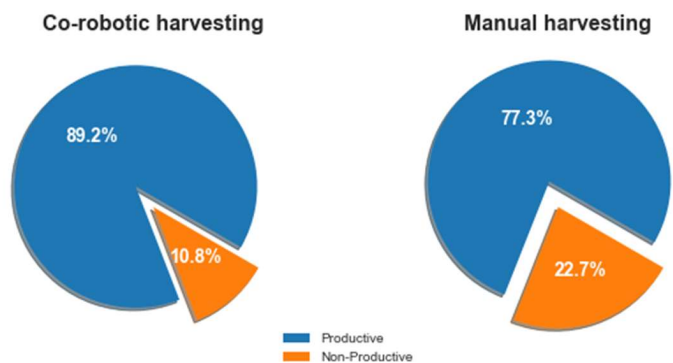

*Figure 26. Comparison between the mean of harvesting efficiency of co-robotic and manual harvesting, based on experimental data, on Nov 10th.*

Similarly, the performance distributions of manual and co-robotic harvesting on Nov 11[th] are shown in Figure 27, and the Mann-Whitney rank test results are shown in Table 10. Based on the calculated P-values, the manual and co-robotic distributions of the non-productive time and efficiency are significantly different, at a significance level of 1%. From Figure 27b, one can see that the mean non-productive time with the robots was 33% lower than that of manual harvesting. Figure 28 shows that the mean harvesting efficiency after introducing the robots improved by 8.8 %.

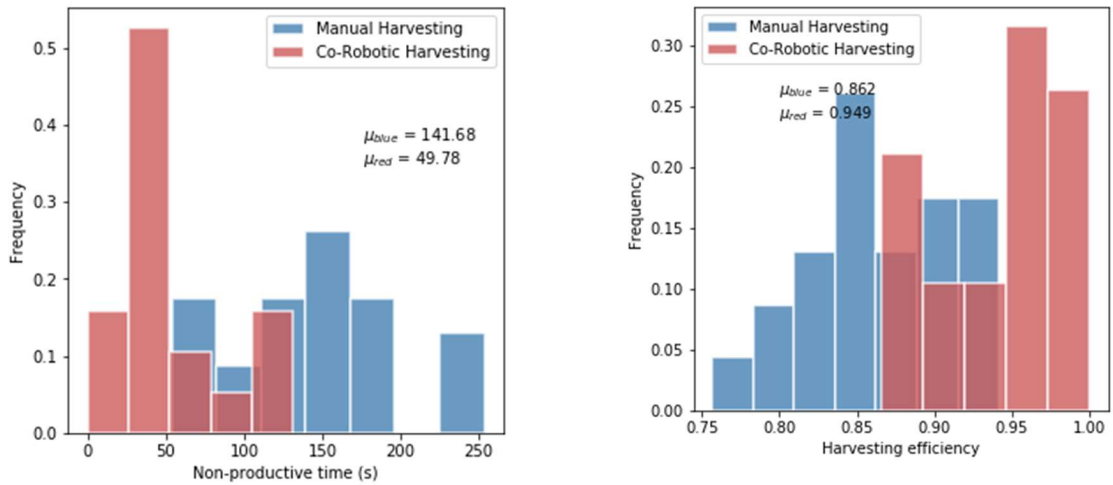

*Figure 27. Harvesting performance on Nov 11[th]: a) Histogram of non-productive time of co-robotic and manual harvesting; b) Histogram of harvesting efficiency of co-robotic and manual harvesting.*

*Table 10 Mann-Whitney rank test results for the measured and estimated non-productive time and harvesting efficiency of the co-robotic and manual harvesting, respectively, on Nov 11[th]*

| Item | Non-productive time | Harvesting efficiency |
|------|---------------------|------------------------|
| T score | 6.253 | 6.073 |
| P value | 2.409e-7 | 3.723e-7 |

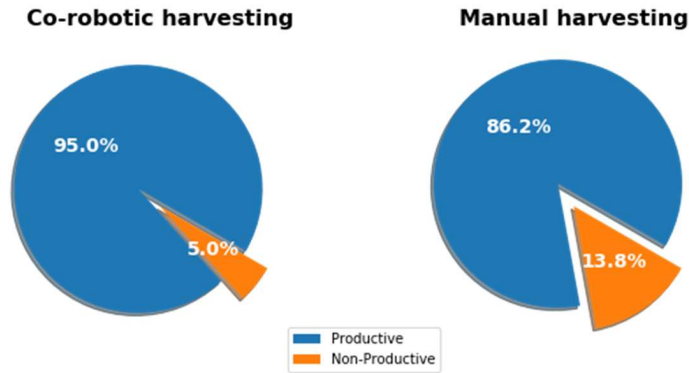

*Figure 28. Comparison between the mean of harvesting efficiency of co-robotic and manual harvesting, based on experimental data, on Nov 11th.*

# 6. Conclusions and future work

This work presented the development of a harvest-aid system with a team of co-robotic crop-transport robots, with strawberry harvesting as a target application. Dynamic predictive robot scheduling was mathematically modeled as a dynamic stochastic scheduling problem with uncertain requests. The modeled problem was solved with a multiple scenario-sampling approach (MSA), which computed near-optimal solutions in real-time. Simulations were used to test the algorithms and select main parameters, and the overall system was deployed during commercial strawberry harvesting.

The system was deployed successfully, and the experimental results showed that the harvest-aid system significantly reduced the non-productive time of the pickers by over 60% and improved the harvesting efficiency by up to 10%, at a robot/picker ratio equal to 1:3.

Future work will focus on two key improvements, which stem from observations and lessons learned during the field experiments. First, the capacity of the current FRAIL-Bot design is more than one tray, and therefore, the robot does not need to return to the unloading station immediately after getting one tray from a picker. Serving multiple pickers before returning to the collection station is expected to increase the efficiency. The scheduling problem when capacity is larger than one is more complex and would need to be solved in real-time. This extension would render the methodology presented in this paper applicable to a wider class of crop-transport in-field logistics applications.

Second, the interaction of human pickers and FRAIL-Bots needs to be improved. For example, the robots were programmed to stop five meters away from the predicted location of the transport request, for safety purposes. Sometimes (e.g., when the fruit load was low) the prediction of the location of the request was too 'aggressive', i.e., the robot stopped too close to the picker. The picker needed to move farther to fill their tray, but the robot was blocking their way. Other times (e.g., when the fruit load was high), the prediction was overly 'conservative', and the picker had to walk a longer distance to the robot. In a couple of instances, we had to stop the robot, because it seemed that it might stop 'behind the picker', i.e., it might collide with them. Better use of existing sensor data and the addition of perception modalities (e.g., onboard cameras) to estimate the operating states of the pickers could enhance human-robot collaboration and safety.